\documentclass[10pt,journal,compsoc, twoside, final]{IEEEtran}

\usepackage{graphicx, caption}
\usepackage{booktabs} 

\usepackage{latexsym}
\usepackage{url}
\usepackage{amsmath,amssymb,amsfonts}
\usepackage{hyperref}
\usepackage[capitalize]{cleveref}
\usepackage{graphicx,bm}
\usepackage[shortlabels]{enumitem}
\usepackage{multirow,multicol}
\usepackage{bigstrut}
\usepackage{MnSymbol}
\usepackage{algorithm}
\usepackage{color,soul}
\usepackage{enumitem}
\usepackage{csquotes}
\usepackage{array}
\usepackage{microtype}
\usepackage{pifont}
\usepackage{subcaption}
\usepackage[round]{natbib} 

\usepackage{cleveref}

\newlength{\bibitemsep}
\newlength{\bibparskip}
\let\oldthebibliography\thebibliography
\renewcommand\thebibliography[1]{%
  \oldthebibliography{#1}%
  \setlength{\parskip}{\bibitemsep}%
  \setlength{\itemsep}{\bibparskip}%
}

\begin{document}
\markboth{}%
{P\MakeLowercase{oria \textit{et al.,}} Beneath the Tip of the Iceberg: Current Challenges and New Directions in Sentiment Analysis Research}



\title{\fontsize{25}{30}\fontfamily{qtm}\selectfont{{Beneath the Tip of the Iceberg: Current Challenges and New Directions in Sentiment Analysis Research}}}




\author{\fontsize{10}{12}\fontfamily{cmr}\selectfont{\textbf{Soujanya Poria$^{\alpha *}$\thanks{$^{*}$ Corresponding author (e-mail: sporia@sutd.edu.sg)}, Devamanyu Hazarika$^{\beta}$, Navonil Majumder$^{\alpha}$, Rada Mihalcea$^{\gamma}$}\\
	
	\vspace{5mm}
	{$^{\alpha}$Information Systems Technology and Design, Singapore University of Technology and Design, Singapore}\\
	{$^{\beta}$ School of Computing, National University of Singapore, Singapore}\\
	{$^{\gamma}$ Electrical Engineering and Computer Science, University of Michigan, Michigan, USA}\\
	\IEEEcompsocitemizethanks{\IEEEcompsocthanksitem Soujanya Poria can be contacted at sporia@sutd.edu.sg
\IEEEcompsocthanksitem Devamanyu Hazarika can be contacted at hazarika@comp.nus.edu.sg
\IEEEcompsocthanksitem Navonil Majumder can be contacted at navonil\_majumder@sutd.edu.sg
\IEEEcompsocthanksitem Rada Mihalcea can be contacted at mihalcea@umich.edu
}

\vspace{7mm}	
\texttt{\textsc{We dedicate this paper to the memory of \textbf{Prof. Janyce Wiebe},\\ who had always believed in the future of the field of sentiment analysis.}}

}
}

\vskip 0.1in
\IEEEtitleabstractindextext{
\begin{abstract}
Sentiment analysis as a field has come a long way since it was first introduced as a task nearly 20 years ago. It has widespread commercial applications in various domains like marketing, risk management, market research, and politics, to name a few. Given its saturation in specific subtasks --- such as sentiment polarity classification --- and datasets, there is an underlying perception that this field has reached its maturity. In this article, we discuss this perception by pointing out the shortcomings and under-explored, yet key aspects of this field necessary to attain \emph{true} sentiment understanding. We analyze the significant leaps responsible for its current relevance. Further, we attempt to chart a possible course for this field that covers many overlooked and unanswered questions.
\end{abstract}

}
\IEEEpeerreviewmaketitle

\maketitle
\IEEEdisplaynontitleabstractindextext
\IEEEraisesectionheading{\section{Introduction}
\label{sec:intro}}
\IEEEPARstart{S}{entiment} analysis, also known as opinion mining, is a research field that aims at understanding the underlying sentiment of unstructured content. E.g.,  in this sentence, ``\textit{John dislikes the camera of iPhone 7}'', according to the technical definition~\citep{liu2012sentiment} of sentiment analysis, John plays the role of the \emph{opinion holder} exposing his \emph{negative sentiment} towards the \emph{aspect} -- camera of the \emph{entity} -- iPhone 7. 
Since its early beginnings~\citep{pang2002thumbs,turney2002thumbs}, sentiment analysis has established itself as an influential field of research with widespread applications in industry. Its ever-increasing popularity and demand stem from the individuals, businesses, and governments interested in understanding people's views about products, political agendas, or marketing campaigns. Public opinion also stimulates market trends, which makes it relevant for financial predictions. Furthermore, education and healthcare sectors make use of sentiment analysis for behavioral analysis of students and patients.


Over the years, the scope for innovation and commercial demand has jointly driven research in sentiment analysis. However, there has been an emerging perception that the problem of sentiment analysis is merely a text/content categorization task -- one that requires content to be classified into two or three categories of sentiments: positive, negative, or neutral. This has led to a belief among researchers that sentiment analysis has reached its saturation. Through this work, we set forth to address this misconception.

\begin{figure*}[t]
	\includegraphics[width=\linewidth]{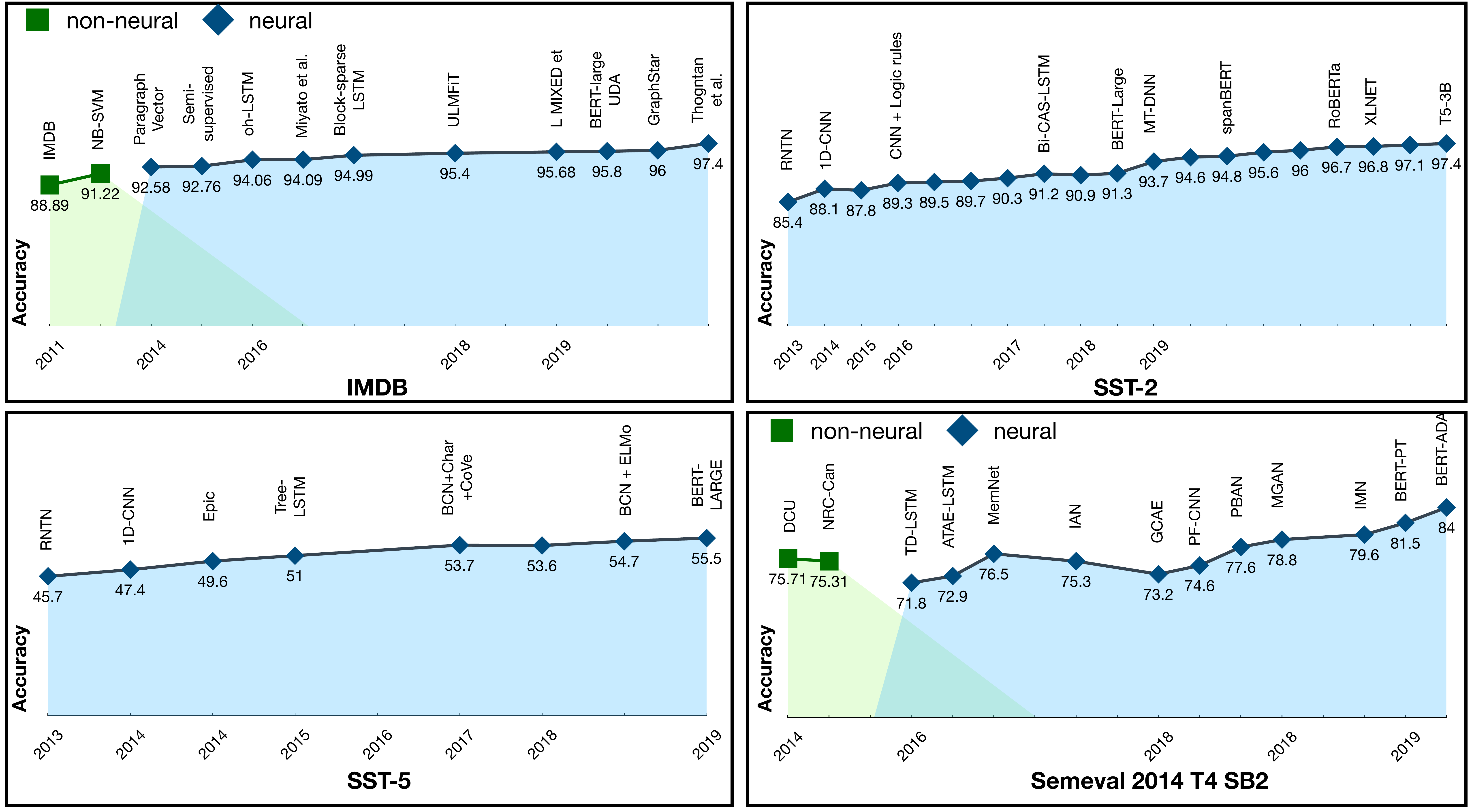}  
	\caption[]{Performance trends of recent models on IMDB~\citep{maas2011learning}, SST-2, SST-5~\citep{socher2013recursive} and Semeval~\citep{DBLP:conf/semeval/PontikiGPPAM14} datasets. The tasks involve sentiment classification in either aspect or sentence level. Note: Data obtained from \url{https://paperswithcode.com/task/sentiment-analysis}. The labels on the graphs are either citations or system names that are also retrieved from \url{https://paperswithcode.com/task/sentiment-analysis}.
	\label{fig:imdb_trends}}
\end{figure*}

\Cref{fig:imdb_trends} shows that many benchmark datasets on the polarity detection subtask of sentiment analysis, like IMDB or SST-2, have reached saturation points, as reflected by the near-perfect scores achieved by many modern data-driven methods. However, this does not imply that sentiment analysis is solved. Rather, we believe that this perception of saturation has manifested from excessive research publications that focus only on shallow sentiment understanding, such as k-way text classification, whilst ignoring other key un- and under-explored problems relevant to this research field.

	

\citet{DBLP:books/daglib/0036864} presents sentiment analysis as \textit{mini-NLP}, given its reliance on topics covering almost the entirety of NLP. Similarly, \citet{DBLP:journals/expert/CambriaPGT17} characterize sentiment analysis as a big suitcase of subtasks and subproblems, involving open syntactic, semantic, and pragmatic problems. As such, there remain several open research directions to be extensively studied, such as understanding motive and cause of sentiment, sentiment dialogue generation, sentiment reasoning, and so on. At its core, effective inference of sentiment requires an understanding of multiple fundamental problems in NLP. These include assigning polarities to aspects, negation handling, resolving co-references, and identifying syntactic dependencies to exploit sentiment flow. The figurative nature of language also influences sentiment analysis, often exploited using linguistic devices, such as sarcasm and irony. This complex composition of multiple tasks makes sentiment analysis a challenging yet interesting research space. 

Figure \ref{fig:imdb_trends} also demonstrates that the methods with a contextual language model as their backbone, much like in other areas of NLP, have dominated these benchmark datasets. Equipped with millions or billions of parameters, transformer-based networks such as BERT~\citep{devlin2018bert}, RoBERTa~\citep{DBLP:journals/corr/abs-1907-11692}, and their variants have pushed the state-of-the-art to new heights. Despite this performance boost, these models are opaque, and their inner-workings are not fully understood. Thus, the question that remains is how far have we progressed since the beginning of sentiment analysis?~\citep{pang2002thumbs}

The importance of lexical, syntactical, and contextual features has been acknowledged numerous times in the past. Recently, due to the advent of the powerful contextualized word embeddings and networks like BERT, we can compute much better representations of such features. Does this entail \emph{true} sentiment understanding? Not likely, as we are far from any significant achievement in \emph{multi-faceted sentiment research}, such as the underlying motivations behind an expressed sentiment, sentiment reasoning, and so on. As members of this research community, we believe that we should strive to move past simple classification as the benchmark of progress and instead direct our efforts towards learning tangible {\it sentiment understanding}. Taking a step in this direction would include analyzing, customizing, and training modern architectures in the context of sentiment, emphasizing fine-grained analysis and exploration of parallel new directions, such as multimodal learning, sentiment reasoning, sentiment-aware natural language generation, and figurative language.

\begin{figure*}[t] 
	\centering 
	\includegraphics[width=\linewidth]{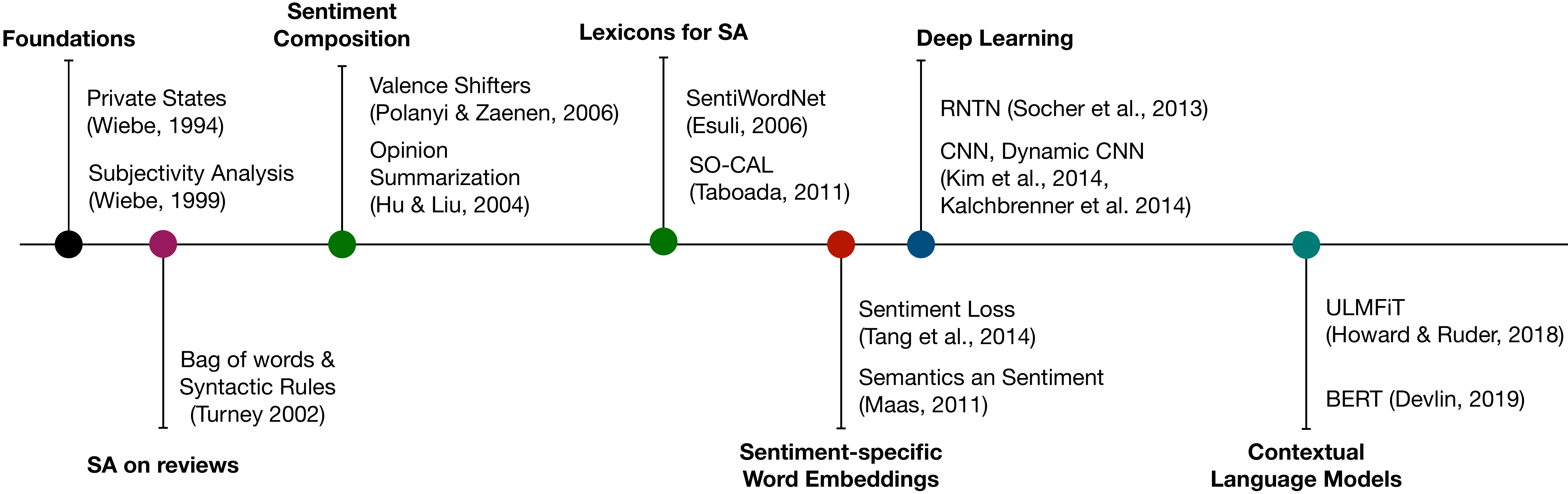} 
	\caption[]{A non-exhaustive illustration of some of the milestones of sentiment analysis research.}
	\label{fig:trends_in_sa}
\end{figure*}

The primary goal of this paper is to motivate new researchers approaching this area. We begin by summarizing the key milestones reached (Figure~\ref{fig:trends_in_sa}) in the last two decades of sentiment analysis research, followed by opening the discussion on new and understudied research areas of sentiment analysis. We also identify some of the critical shortcomings in several sub-fields of sentiment analysis and describe potential research directions. This paper is not intended as a survey of the field -- we mainly cover a small number of key contributions that have either had a seminal impact on this field or have the potential to open new avenues. Our intention, thus, is to draw attention to key research topics within the broad field of sentiment analysis and identify critical directions left to be explored. We also uncover promising new frameworks and applications that may drive sentiment analysis research in the near future.

\begin{figure*}[t] 
	\centering 
	\begin{subfigure}[t]{\textwidth}
	\includegraphics[width=\linewidth]{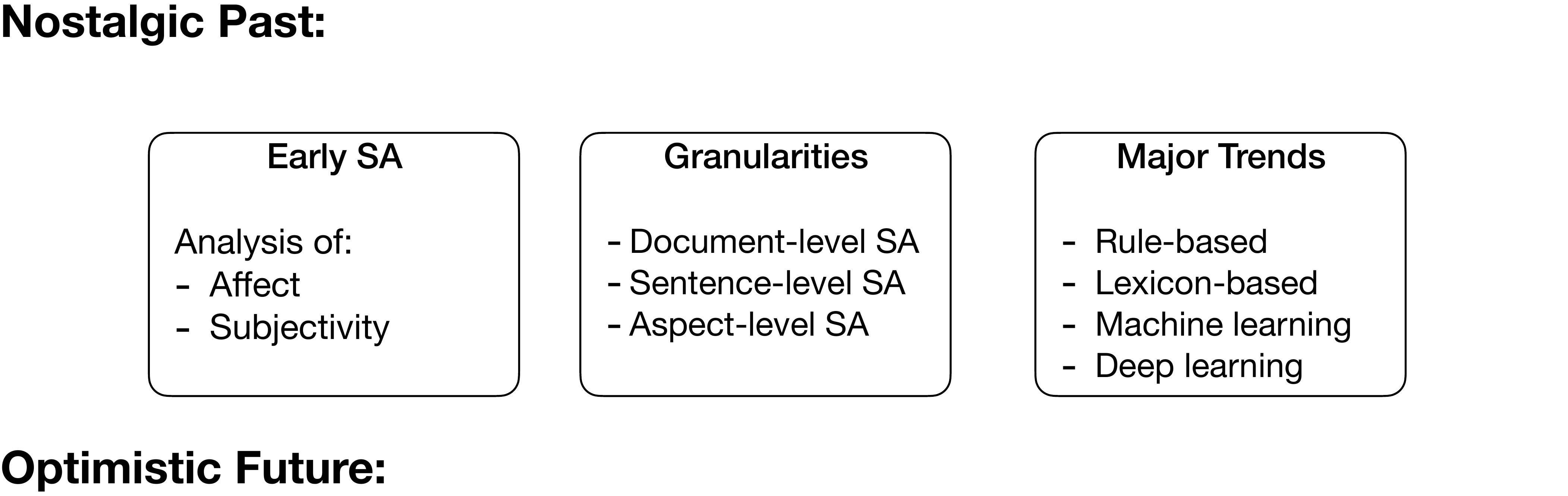} 
	\end{subfigure}
	~
	\begin{subfigure}[t]{\textwidth}
	\includegraphics[width=\linewidth]{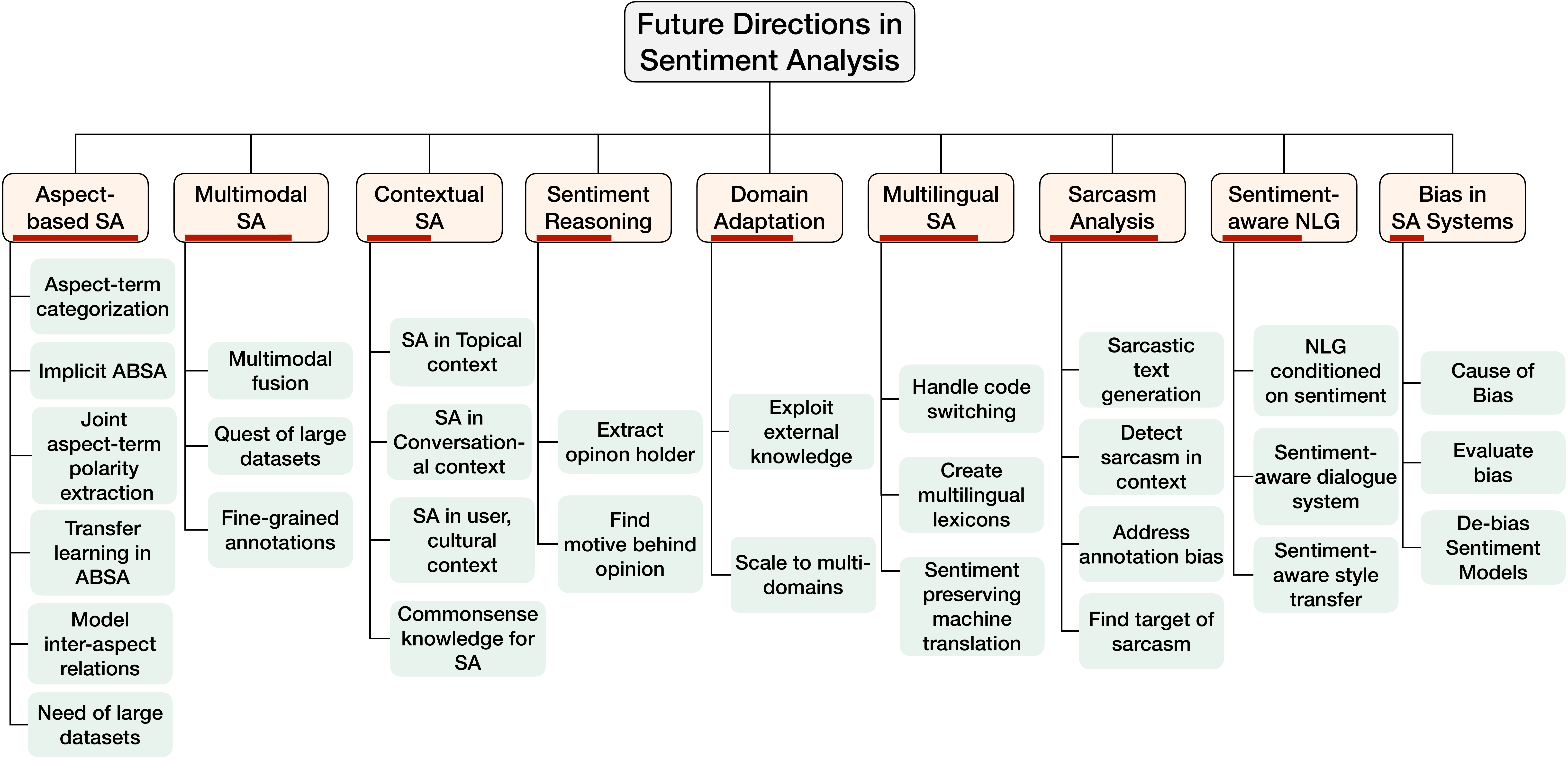} 
	\end{subfigure}
	\caption[]{The paper is logically divided into two sections. First, we analyze the past trends and where we stand today in the sentiment analysis Literature. Next, we present an \textit{Optimistic} peek into the future of sentiment analysis, where we discuss several applications and possible new directions. The red bars in the figure estimates the present popularity of each application. The lengths of these bars are proportional to the logarithm of the publication counts on the corresponding topics in Google Scholar since 2000. Note: SA and ABSA are the acronyms for Sentiment Analysis and Aspect-Based Sentiment Analysis.
	}
	\label{fig:structure}
\end{figure*}

The rest of the paper is organized as follows: Section~\ref{sec:milestones} briefly describes the key developments and achievements in the sentiment analysis research; we discuss the future directions of sentiment analysis research in Section~\ref{sec:future}; and finally, Section~\ref{sec:conclusion} concludes the paper. We illustrate the overall organization of the paper in Figure~\ref{fig:structure}. We curate all the articles, that cover the past and future of sentiment analysis (see Figure~\ref{fig:structure}), on this repository: \url{https://github.com/declare-lab/awesome-sentiment-analysis}.

\section{Nostalgic Past: Developments and Achievements in Sentiment Analysis}
\label{sec:milestones}
The fields of \textbf{sentiment analysis} and \textbf{opinion mining} --- often used as synonyms --- aim to determine the sentiment polarity of unstructured content in the form of text, audio streams, or multimedia-videos.

\subsection{Early Sentiment Analysis} 
 
The task of sentiment analysis originated from the analysis of subjectivity in sentences~\citep{wiebe1999development, DBLP:conf/aaai/Wiebe00, hatzivassiloglou2000effects, yu2003towards, wilson2005opinionfinder}. \citet{wiebe1994tracking} associated subjective sentences with \textit{private states} of the speaker, that are not open for observation or verification, taking various forms such as opinions or beliefs. Research in sentiment analysis, however, became an active area only since 2000 primarily due to the availability of opinionated online resources \citep{tong2001operational, morinaga2002mining, nasukawa2003sentiment}. One of the seminal works in sentiment analysis involves categorizing reviews based on their orientation (sentiment)~\citep{turney2002thumbs}. This work generalized phrase-level orientation mining by enlisting several syntactic rules~\citep{hatzivassiloglou1997predicting} and also introduced the bag-of-words concept for sentiment labeling. It stands as one of the early milestones in developing this field of research.

Although preceded by related tasks, such as identifying affect, the onset of the $21^{st}$ century marked the surge of modern-day sentiment analysis.

\subsection{Granularities} 

Traditionally, sentiment analysis research has mainly focused on three levels of granularity~\citep{liu2012sentiment,liu2010sentiment}: document-level, sentence-level, and aspect-level sentiment analysis.

In \textit{\textbf{document-level sentiment analysis}}, the goal is to infer the overall opinion of a document, which is assumed to convey a unique opinion towards an entity, e.g., a product~\citep{pang2004sentimental, glorot2011domain, moraes2013document}. \citet{pang2002thumbs} conducted one of the initial works on document-level sentiment analysis, where they assigned positive/negative polarity to review documents. They used various features, including unigrams (bag of words) and trained simple classifiers, such as Naive Bayes classifiers and SVMs. Although primarily framed as a classification/regression task, alternate forms of document-level sentiment analysis research include other tasks such as generating opinion summaries~\citep{DBLP:conf/aaaiss/KuLC06,DBLP:conf/naacl/LloretBPM09}.

\textit{\textbf{Sentence-level sentiment analysis}} restricts the analysis to individual sentences~\citep{yu2003towards, kim2004determining}. These sentences could belong to documents, conversations, or standalone micro-texts found in resources such as microblogs~\citep{kouloumpis2011twitter}. 

While both document- and sentence-level sentiment analysis provide an overall sentiment orientation, they do not indicate the target of the sentiment in many cases. They implicitly assume that the text span (document or sentence) conveys a single sentiment towards an entity, which typically is a very strong assumption. 

To overcome this challenge, the analysis is directed towards a finer level of scrutiny, i.e., \textit{\textbf{aspect-level sentiment analysis}}, where sentiment is identified for each entity~\citep{hu2004mining} (along with its aspects). Aspect-level analysis allows a better understanding of the sentiment distribution. We discuss its challenges in Section~\ref{sec:absa}.

In addition to these three granularity levels, a significant amount of studies have been done for \textit{\textbf{phrase-level sentiment analysis}}, which focus on phrases within a sentence~\citep{DBLP:conf/naacl/WilsonWH05}. In this granularity, the goal is to analyze how the sentiment of words, present in lexicons, can change in and out of context, e.g., \textit{good} vs. \textit{doesn't look good}. Many compositional factors, such as negators, modals, or intensifiers, may flip or change the degree of sentiment~\citep{DBLP:conf/semeval/KiritchenkoMS16}, which makes it a difficult yet important task. 

In the initial developments of sentiment analysis, phrase-level study contributed numerous advances in defining rules that accounted for these compositions. We discuss these in the next section. Phrase-level sentiment analysis has also been important in small text pieces found in micro-blogs, such as Twitter. Many recent shared-tasks discuss this area of research~\citep{DBLP:conf/semeval/NakovRKSRW13,DBLP:conf/semeval/RosenthalRNS14,DBLP:conf/semeval/RosenthalNKMRS15} (see~\cref{sec:microblogs}).

\subsection{Trends in Sentiment Analysis Applications} \label{sec:sentiment_trends}

\paragraph*{\textbf{Rule-Based Sentiment Analysis}}

A major section of the history of sentiment analysis research has focused on utilizing sentiment-bearing words and utilizing their compositions to analyze phrasal units for polarity. Early work identified that the simple counting of valence words, i.e., a bag-of-words approach, can provide incorrect results~\citep{polanyi2006contextual}. This led to the emergence of research on \textit{valence shifters} that incorporated changes in valence and polarity of terms based on contextual usage~\citep{polanyi2006contextual, moilanen2007sentiment}. However, only valence shifters were not enough to detect sentiment -- it also required understanding sentiment flows across syntactic units. Thus, researchers introduced the concept of modeling sentiment composition, learned via heuristics and rules~\citep{DBLP:conf/emnlp/ChoiC08}, hybrid systems~\citep{rentoumi2010united}, syntactic dependencies~\citep{nakagawa2010dependency, hutto2014vader}, amongst others.

\textbf{Sentiment Lexicons} are at the heart of rule-based sentiment analysis methods. Simply defined, these lexicons are dictionaries that contain sentiment annotations for their constituent words, phrases, or synsets~\citep{joshi2017sentiment,DBLP:conf/cikm/CambriaLXPK20}.

SentiWordNet~\citep{esuli2007sentiwordnet} is one such popular sentiment lexicon that builds on top of Wordnet~\citep{fellbaum2012wordnet}. In this lexicon, each synset is assigned with positive, negative, and objective scores, which indicate their subjectivity orientation. As the labeling is associated with synsets, the subjectivity score is tied to word senses. This trait is desirable as subjectivity and word-senses have strong semantic dependence, as highlighted in \citet{DBLP:conf/acl/WiebeM06}.

SO-CAL~\citep{DBLP:journals/coling/TaboadaBTVS11}, as the name suggests, presents a lexicon-based sentiment calculator. It contains a dictionary of words associated with their semantic orientation (polarity and strength). The strength of this resource is in its ability to account for contextual valence shifts, which include factors that affect the polarity of words through intensification, lessening, and negations.

Other popular lexicons include SCL-OPP~\citep{DBLP:conf/lrec/KiritchenkoM16}, SCL-NMA~\citep{DBLP:conf/wassa/KiritchenkoM16}, 
amongst others. These lexicons not just store word-polarity associations but also include phrases or rules that reflect complex sentiment compositions, e.g., negations, intensifiers. NRC Lexicon~\cite{mohammad-turney-2010-emotions,DBLP:journals/ci/MohammadT13} is another popular lexicon that hosts both word-sentiment and word-emotion pairs with high-quality annotations.

Lexicons have been mainly created using three broad approaches, manual, automatic, and semi-supervised. In manual creation, expert-annotation~\citep{DBLP:journals/coling/TaboadaBTVS11} or crowd-sourcing~\citep{socher2013recursive} is used to create annotations. In automatic methods, annotations are generated or expanded using knowledge bases, such as Wordnet. The advantage of automatic methods is a broader coverage of instances and also sense-specific annotations, especially for Wordnet-based expansions~\citep{esuli2007sentiwordnet}. However, automatic methods tend to have higher noise in the annotations. The final approach is to use semi-supervised learning, such as label propagation~\citep{zhur2002learning} or graph propagation~\citep{DBLP:conf/naacl/VelikovichBHM10}, to create new sentiment lexicons that could be domain-specific~\citep{DBLP:conf/iiwas/TaiK13} or in new languages~\citep{DBLP:conf/acl/ChenS14} -- two topics that hold high relevance in present research directions.

Though lexicons provide valuable resources for archiving sentiment polarity of words or phrases, utilizing them to infer sentence-level polarities has been quite challenging. Moreover, no one lexicon can handle all the nuances observed from semantic compositionality~\citep{DBLP:conf/coling/Toledo-RonenBHJ18,DBLP:conf/naacl/KiritchenkoM16a} or account for contextual polarity. Lexicons also have many challenges in their creation, such as combating subjectivity in annotations~\citep{mohammad2017challenges}.

While sentiment lexicons remain an integral component of sentiment analysis systems, especially for low-resource instances, there has been an increase in focus towards statistical solutions. These solutions do not suffer from the issue of rules-coverage and provide better opportunities to handle generalization. 

\paragraph*{\textbf{Machine Learning-Based Sentiment Analysis}}
Statistical approaches that employ machine learning have been appealing to this area, particularly due to their independence over hand-engineered rules. Despite best efforts, the rules could never be enumerated exhaustively, which always kept the generalization capability limited. With machine learning, the opportunity to learn generic representations emerged. Throughout the development of sentiment analysis, ML-based approaches--both supervised and unsupervised--have employed myriad of algorithms that include SVMs~\citep{DBLP:journals/eswa/MoraesVN13}, Naive Bayes Classifiers~\citep{DBLP:conf/ecir/TanCWX09}, nearest neighbour~\citep{DBLP:conf/cikm/MoghaddamE10}, combined with features that range from bag-of-words (including weighted variants)~\citep{DBLP:conf/icwsm/MartineauF09}, lexicons~\citep{DBLP:journals/eswa/GavilanesAJCG16} to syntactic features such as parts of speech~\citep{DBLP:conf/icwsm/MejovaS11}. 
A detailed review for most of these works has been provided in  \citep{liu2010sentiment, liu2012sentiment}.

\paragraph*{\textbf{Deep Learning Era}}
The advent of deep learning saw the use of distributional embeddings and techniques for representation learning for various tasks of sentiment analysis. One of the initial models was the \textit{Recursive Neural Tensor Network (RNTN)}~\citet{socher2013recursive}, which determined the sentiment of a sentence by modeling the compositional effects of sentiment in its phrases. This work also introduced the \textit{Stanford Sentiment Treebank} corpus comprising of parse trees fully labeled with sentiment labels. The unique usage of recursive neural networks adapted to model the compositional structure in syntactic trees was highly innovative and influencing~\citep{DBLP:conf/acl/TaiSM15}.

CNNs and RNNs were also used for feature extraction. The popularity of these networks, especially that of CNNs, can be traced back to \citet{kim2014convolutional}. Although CNNs had been used in NLP systems earlier~\citep{collobert2011natural}, the investigatory work by~\citet{kim2014convolutional} presented a CNN architecture which was simple (single-layered) and also delved into the notion of non-static embeddings. It was a popular network, that became the de-facto sentential feature extractor for many of the sentiment analysis tasks. Similar to CNNs, RNNs also enjoyed high popularity. Aside from polarity prediction, these architectures outperformed traditional graphical models in structured prediction tasks such as aspect, aspect-term and opinion-term extraction~\citep{DBLP:journals/kbs/PoriaCG16,DBLP:conf/emnlp/IrsoyC14}. Aspect-level sentiment analysis, in particular, saw an increase in complex neural architectures that involve attention mechanisms~\citep{wang2016attention}, memory networks~\citep{DBLP:conf/emnlp/TangQL16} and adversarial learning~\citep{DBLP:journals/corr/abs-2001-11316,DBLP:journals/tacl/ChenSACW18}. For a comprehensive review of modern deep learning architectures, please refer to~\citep{zhang2018deep}.

Although the majority of the works employing deep networks rely on automated feature learning, their heavy reliance on annotated data is often limiting. As a result, providing inductive biases via syntactic information, or external knowledge in the form of lexicons as additional input has seen a resurgence~\citep{DBLP:conf/emnlp/TayLHS18}.

As seen in Figure~\ref{fig:imdb_trends}, the recent works based on neural architectures~\citep{le2014distributed,dai2015semi,johnson2016supervised,miyato2016adversarial,mccann2017learned,howard2018universal,xie2019unsupervised,DBLP:conf/acl/ThongtanP19} have dominated over traditional machine learning models~\citep{maas2011learning,wang2012baselines}. Similar trends can be observed in other benchmark datasets such as Yelp, SST~\citep{socher2013recursive}, and Amazon Reviews~\citep{DBLP:conf/nips/ZhangZL15}. Within neural methods, much like other fields of NLP, present trends are dominated by the contextual encoders, which are pre-trained as language models using the Transformer architecture~\citep{DBLP:conf/nips/VaswaniSPUJGKP17}. Models like BERT, XLNet, RoBERTa, and their adaptations have achieved the state-of-the-art performances on multiple sentiment analysis datasets and benchmarks~\citep{DBLP:conf/nodalida/HoangBR19,DBLP:journals/corr/abs-1910-03474,DBLP:journals/corr/abs-1910-10683}. Despite this progress, it is still not clear as to whether these new models learn the composition semantics associated to sentiment or simply learn surface patterns~\citep{DBLP:journals/corr/abs-2002-12327}.

\paragraph*{\textbf{Sentiment-Aware Word Embeddings}}
One of the critical building blocks of a text-processing deep-learning architecture is its word embeddings. It is known that the word representations rely on the task it is being used for~\citep{labutov2013re}. However, most sentiment analysis-based models use static general-purpose word representations. \citet{tang2014learning} proposed an important work in this direction that provided word representations tailored for sentiment analysis. While general embeddings mapped words with similar syntactic context into nearby representations, this work incorporated sentiment information into the learning loss to account for sentiment regularities. Although the community has proposed some approaches in this topic~\citep{maas2011learning,DBLP:conf/cikm/BespalovBQS11},
promising traction has been limited~\citep{tang2015document}. Further, with the popularity of contextual models such as BERT, it remains to be seen how sentiment information can be incorporated into its embeddings.

\begin{figure*}[ht!] 
    \centering 
    \includegraphics[width=\linewidth]{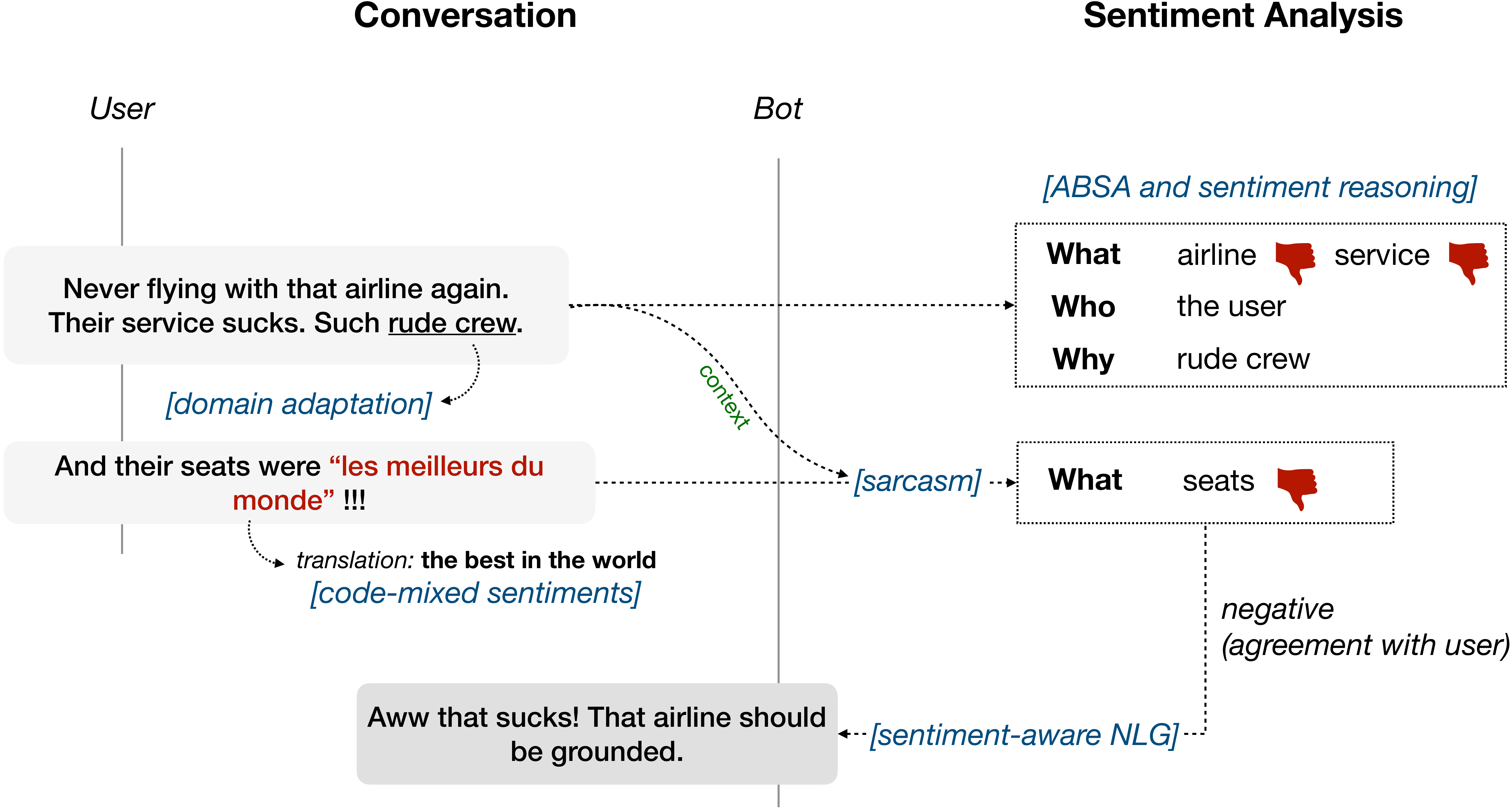} 
    \caption[]{The example illustrates the various challenges and applications that holistic sentiment analysis depends on.}
    \label{fig:sentiment-flowchart}
\end{figure*}

\subsection{Sentiment Analysis in Diverse Domains} \label{sec:microblogs}

Sentiment analysis in micro-blogs, such as Twitter, require different processing techniques compared to traditional text pieces. Enforced maximum text length often coerces users to express their opinion straightforwardly. However, sarcasm and irony pose a challenge to these systems. Tweets are rife with internal slangs, abbreviations, and emoticons -- which adds to the complexity of mining their opinions. Moreover, the limited length restricts the presence of contextual cues normally present in dialogues or documents~\citep{DBLP:journals/corr/KhardeS16}.

From a data point of view, opinionated data is found in abundance in such micro-blogs. Reflections of this have been observed in the recent benchmark shared tasks based on Twitter data. These include Semeval shared tasks for sentiment analysis, aspect-based sentiment analysis and figurative language in Twitter~\footnote{\url{http://alt.qcri.org/semeval2015/task10/}}\textsuperscript{,}~\footnote{\url{http://alt.qcri.org/semeval2015/task12/}}\textsuperscript{,}~\footnote{\url{http://alt.qcri.org/semeval2015/task11/}}\textsuperscript{,}~\footnote{\url{https://sites.google.com/view/figlang2020/}}.

A new trend amongst users in Twitter is the concept of daisy-chaining multiple tweets to compose a longer piece of text. Existing research, however, has not addressed this phenomenon to acquire additional context.
Future work on Twitter sentiment analysis could benefit from analyzing users' personalities based on their historical tweets.

The application of sentiment analysis is not limited to social media and review articles. It spans from emails~\citep{9004372} to financial documents~\citep{Krishnamoorthy_2017} showing the efficacy in understanding the mood of users across different domains.

\section{Optimistic Future: Upcoming Trends in sentiment analysis}
\label{sec:future}

The previous section highlighted some of the milestones in sentiment analysis research, which helped develop the field into its present state. Despite the progress, we believe the problems are far from solved. In this section, we take an optimistic view on the road ahead in sentiment analysis research and highlight several applications rife with open problems and challenges. 

Applications of sentiment analysis take form in many ways.~\cref{fig:sentiment-flowchart} presents one such example where a user is chatting with a chit-chat style chatbot. In the conversation, to generate an appropriate response, the bot needs to understand the user's opinion. This involves multiple sub-tasks that include 1) extracting aspects like service, seats for the entity airline, 2) aspect-level sentiment analysis along with knowing 3) \textit{who} holds the opinion and \textit{why} (sentiment reasoning). Added challenges include analyzing code-mixed data (e.g. ``\textit{les meilleurs du monde}''), understanding domain-specific terms (e.g., rude crew), and handling sarcasm -- which could be highly contextual and detectable only when preceding utterances are taken into consideration. Once the utterances are understood, the bot can now determine appropriate response-styles and perform controlled-natural language generation (NLG) based on the decided sentiment. The overall example demonstrates the dependence of sentiment analysis on these applications and sub-tasks, some of which are new and still at early development stages. We discuss these applications next. 

\subsection{Aspect-Based Sentiment Analysis} \label{sec:absa}

Although sentiment analysis provides an overall indication of the author or speaker's sentiments, it is often the case when a piece of text comprises multiple aspects with varied sentiments. For example, in the following sentence, ``\emph{This actor is the only failure in an otherwise brilliant cast.}'', the opinion is attached to two particular entities, \textit{actor} (negative opinion) and \textit{cast} (positive opinion). There is also an absence of an overall opinion that could be assigned to the full sentence.

\textit{Aspect-based Sentiment Analysis} (ABSA) takes such fine-grained view and aims to identify the sentiments towards each entity (or their aspects)~\citep{DBLP:books/daglib/0036864,DBLP:books/sp/mining2012/LiuZ12}. The problem involves two major sub-tasks, 1) \textit{Aspect-extraction}, which identifies the aspects~\footnote{In the context of aspect-based sentiment analysis, \textit{aspect} is the generic term utilized for topics, entities, or their attributes/features. They are also known as opinion targets.} mentioned within a given sentence or paragraph (\textit{actor} and \textit{cast} in the above example) 2) \textit{Aspect-level Sentiment Analysis} (ALSA), which determines the sentiment orientation associated with the corresponding aspects/ opinion targets (actor $\mapsto$ negative and cast $\mapsto$ positive)~\citep{DBLP:conf/kdd/HuL04}. Proposed approaches for aspect extraction include rule-based strategies~\citep{DBLP:journals/coling/QiuLBC11,DBLP:conf/ijcai/LiuGLZ15}, topic models~\citep{DBLP:conf/www/MeiLWSZ07, DBLP:conf/acl/HeLA11}, and more recently, sequential structured-prediction models such as CRFs~\citep{DBLP:conf/acl/ShuXL17}. For aspect-level sentiment analysis, the algorithms primarily aim to model the relationship between the opinion targets and their context. To achieve this, models based on CNNs~\citep{DBLP:conf/aaai/LiL17}, memory networks~\citep{DBLP:conf/cikm/TayTH17}, etc. have been explored. Primarily, the associations have been learnt through attention mechanism~\citep{wang2016attention}.

Despite the advances in this field, there remain many factors which are open for research and hold the potential to improve performances further. We discuss them below.

\begin{figure}[t] 
    \centering 
    \includegraphics[width=\linewidth]{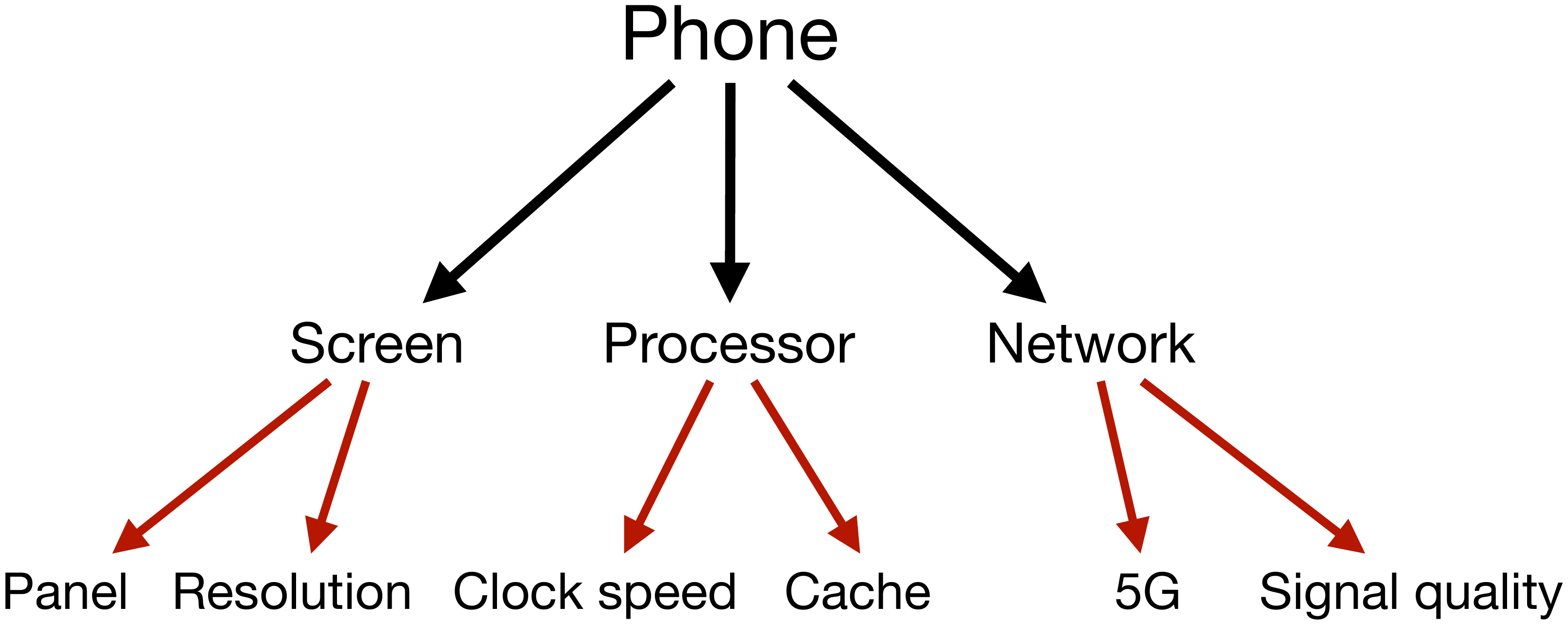} 
    \caption[]{An example of the aspect term auto-categorization.}
    \label{fig:aspect-hierarchy}
\end{figure}

\subsubsection{Aspect-Term Auto-Categorization}
Aspect-terms extraction is the first step towards aspect-level sentiment analysis. This task has been studied rigorously in the literature~\citep{DBLP:journals/kbs/PoriaCG16}. Thanks to the advent of deep sequential learning, the performance of this task on the benchmark datasets~\citep{hu2004mining,pontiki2016semeval} has reached a new level. 
Aspect terms are needed to be categorized into aspect groups to present a coherent view of the expressed opinions. We illustrate this categorization in \cref{fig:aspect-hierarchy}. Approaches to aspect-term auto-categorization are mostly based on supervised and unsupervised topic classification and lexicon driven. These three types of approaches succumb to scalability issues when subjected to new domains with novel aspect categories. We believe that entity linking-based approaches, coupled with semantic graphs like Probase~\citep{probase}, should be able to perform reasonably while overcoming scalability issues. For example, the sentence ``\emph{With this phone, I always have a hard time getting \ul{signal} indoors.}'' contains one aspect term \emph{signal}, that can be passed to an entity linker --- on a graph containing a tree shown in~\cref{fig:aspect-hierarchy} --- with the surrounding words as context to obtain aspect category \emph{phone:signal-quality}.
\subsubsection{Implicit Aspect-Level Sentiment Analysis}
Sentiment on different aspects can be expressed implicitly. 
Although under-studied, the importance of detecting implicit aspect-level sentiment can not be ignored as they represent a unique nature of natural language. 
For example, in the sentence,
``\textit{Oh no! Crazy Republicans voted against this bill}", the speaker explicitly expresses her/his negative sentiment on the \emph{Republicans}. At the same time, we can infer that the speaker's sentiment towards the \emph{bill} is positive. In the work by \citet{deng2014joint}, it is called as \emph{opinion-oriented implicatures}. As most datasets are not labeled with such details, models trained on present datasets might miss extracting \emph{bill} as an aspect term and its associated polarity. Thus, this remains an open problem in the overall ABSA task. The case of implicit sentiment is also observed in contextual sentiment analysis, which we further discuss in~\cref{sec:sa_in_conversations}.

\subsubsection{Aspect Term-Polarity Co-Extraction}

Most existing algorithms in this area consider aspect extraction and aspect-level sentiment analysis as sequential (pipelined) or independent tasks. In both these cases, the relationship between the tasks is ignored. Efforts towards joint learning of these tasks have gained traction in recent trends. These include hierarchical neural networks \citep{lakkaraju2014aspect}, multi-task CNNs \citep{DBLP:conf/ijcnn/WuGSG16}, and CRF-based approaches by framing both the sub-tasks as sequence labeling problems~\citep{DBLP:conf/aaai/LiBLL19,DBLP:conf/acl/LuoLLZ19}. The notion of joint learning opens up several avenues for exploring the relationships between the sub-tasks and possible dependencies from other tasks. 

Another strategy is to leverage transfer learning since aspect extraction can be utilized as a scaffolding for aspect-based sentiment analysis~\citep{DBLP:journals/corr/abs-2005-06607}. Knowledge transfer can also be observed from textual to multimodal ABSA system.




\subsubsection{Exploiting Inter-Aspect Relations for Aspect-Level Sentiment Analysis}

The primary focus of algorithms proposed for aspect-level sentiment analysis has been to model the dependencies between opinion targets and their corresponding opinionated words in the context~\citep{DBLP:conf/coling/TangQFL16}. Besides, modeling the relationships between aspects also holds potential in this task \citep{DBLP:conf/naacl/HazarikaPVKCZ18}. For example, in the sentence "my favs here are the tacos pastor and the tostada de tinga",  the aspects
\emph{"tacos pastor"} and \emph{"tostada de tinga"} are connected using conjunction \emph{"and"} and both rely on the sentiment bearing word \emph{"favs"}. Understanding such inter-aspect dependency can significantly aid the aspect-level sentiment analysis performance and remains to be researched extensively.

\subsubsection{Quest for Richer and Larger Datasets}
The two widely used publicly available datasets for aspect-based sentiment analysis are Amazon product review~\citep{hu2004mining} and Semeval-2016~\citep{pontiki2016semeval} datasets. Both these datasets are quite small in size that hinders any statistically significant performance improvement between methods utilizing them. 



\subsection{Multimodal Sentiment Analysis} \label{sec:multimodal_sa}

\begin{figure}[t] 
    \centering 
    \includegraphics[width=\linewidth]{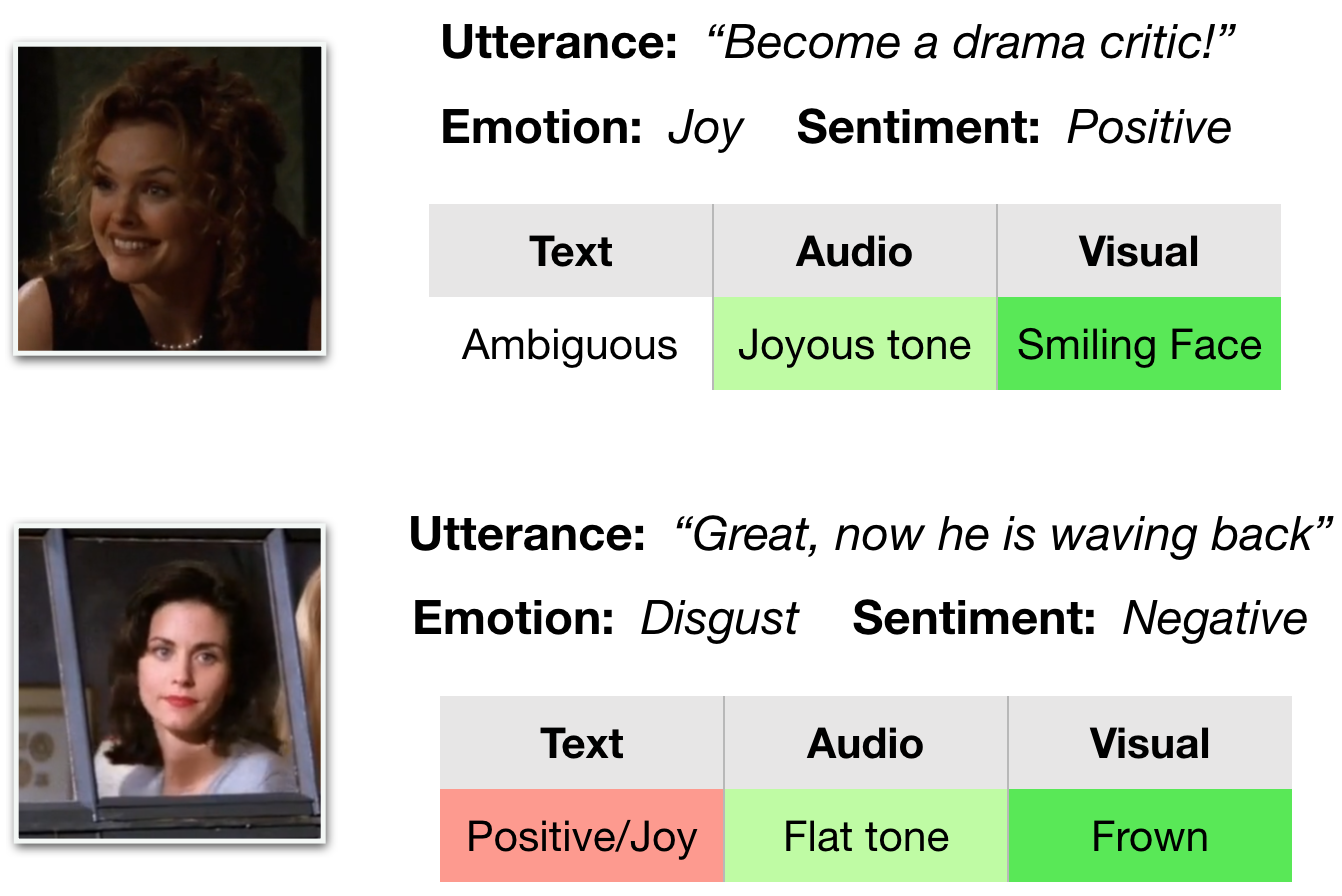} 
    \caption[]{Importance of multimodal cues. Green shows primary modalities responsible for sentiment and emotion.}
    \label{fig:multimodal_examples}
\end{figure}

The majority of research works on sentiment analysis have been conducted using only textual modality. However, with the increasing number of user-generated videos available on online platforms such as YouTube, Facebook, Vimeo, and others, multimodal sentiment analysis has emerged at the forefront of sentiment analysis research. The commercial interests fuel this rise as the enterprises tend to make business decisions on their products by analyzing user sentiments in these videos. Figure~\ref{fig:multimodal_examples} presents examples where the presence of multimodal signals in addition to the text itself is necessary in order to make correct predictions of their emotions and sentiments. Multimodal fusion is at the heart of multimodal sentiment analysis, with an increasing number of works proposing new fusion techniques. These include Multiple Kernel Learning, tensor-based non-linear fusion~\citep{zadeh2017tensor}, memory networks~\citep{zadmem}, amongst others. The granularity at which such fusion methods are applied also varies -- from word-level to utterance-level.

Below, we identify three key directions that can aid future research:

\subsubsection{Complex Fusion Methods vs. Simple Concatenation}
Multimodal information fusion is a core component of multimodal sentiment analysis. Although several fusion techniques~\citep{zadatt,zadmem,zadeh2017tensor} have been recently proposed, in our experience, a simple concatenation-based fusion method performs at par with most of these methods. We believe these methods are unable to provide significant improvements in the fusion due to their inability to model correlations among different modalities and handle noise. Reliable fusion remains a major future work.

\subsubsection{Lack of Large Datasets}
The field of multimodal sentiment analysis also suffers from the lack of larger datasets. The available datasets, such as MOSI~\citep{DBLP:journals/expert/ZadehZPM16}, MOSEI~\citep{DBLP:conf/acl/MorencyCPLZ18}, MELD~\citep{DBLP:conf/acl/PoriaHMNCM19} are not large enough and carry suboptimal inter-annotator agreement that impedes the performance of complex deep learning frameworks.

\subsubsection{Fine-Grained Annotation}
The primary goal of multimodal fusion is to accumulate the contribution from each modality. However, measuring that contribution is not trivial as there is no available dataset that annotates the individual role of each modality. We show one such example in Figure~\ref{fig:multimodal_examples}, where each modality is labeled with the sentiment it carries.  Having such rich fine-grained annotations should better guide multimodal fusion methods and make them more interpretable. This fine-grained annotation can also open the door to novel multimodal fusion approaches.

\subsection{Contextual Sentiment Analysis} \label{sec:contextual_sa}

\subsubsection{Influence of Topics}
The usage of sentiment words varies from one topic to another. Words that sound neutral on the surface can bear sentiment when conjugated with other words or phrases. For example, the word \emph{big} in \emph{big house} can carry positive sentiment when someone intends to purchase a \emph{big house} for leisure. However, the same word could evoke negative sentiments when used in the context -- \emph{A big house is hard to clean}. Unfortunately, research in sentiment analysis has not focused much on this aspect. The sentiment of some words can be vague and specified only when seen in context, e.g., the word \emph{massive} in the context of \emph{massive earthquake} and \emph{massive villa}.  A dataset composed of such contextual sentiment bearing phrases would be a great contribution to the research community in the future. 

This research problem is also related to word sense disambiguation. Below we present an example, borrowed from the work by \citet{choi2017coarse}:
\begin{enumerate}[a.]
\item The Federal Government \emph{carried} the province for many years.
\item The troops \emph{carried} the town after a brief fight.
\end{enumerate}

In the first sentence, the sense of \emph{carry} has a positive polarity. However, in the second sentence, the same word has negative connotations. Hence, depending on the context, the sense of words and their polarities can change. In \cite{choi2017coarse}, the authors adopted topic models to associate word senses with sentiments. As this particular research problem widens its scope to the task of word sense disambiguation, it would be useful to employ contextual language models to decipher word senses in contexts and assign the corresponding polarity.

\subsubsection{Sentiment Analysis in Monologues and Conversational Context} 
\label{sec:sa_in_conversations}

Context is at the core of NLP research. According to several recent studies~\citep{peters2018deep,devlin2018bert}, contextual sentence and word-embeddings can improve the performance of the state-of-the-art NLP systems by a significant margin.

The notion of context can vary from problem to problem. For example, while calculating word representations, the surrounding words carry contextual information. Likewise, to classify a sentence in a document, other neighboring sentences are considered as its context. \citet{poria2017context} utilize surrounding utterances in a video as context and experimentally show that contextual evidence indeed aids in classification.

\begin{figure}[t] 
    \centering 
    \includegraphics[width=\linewidth]{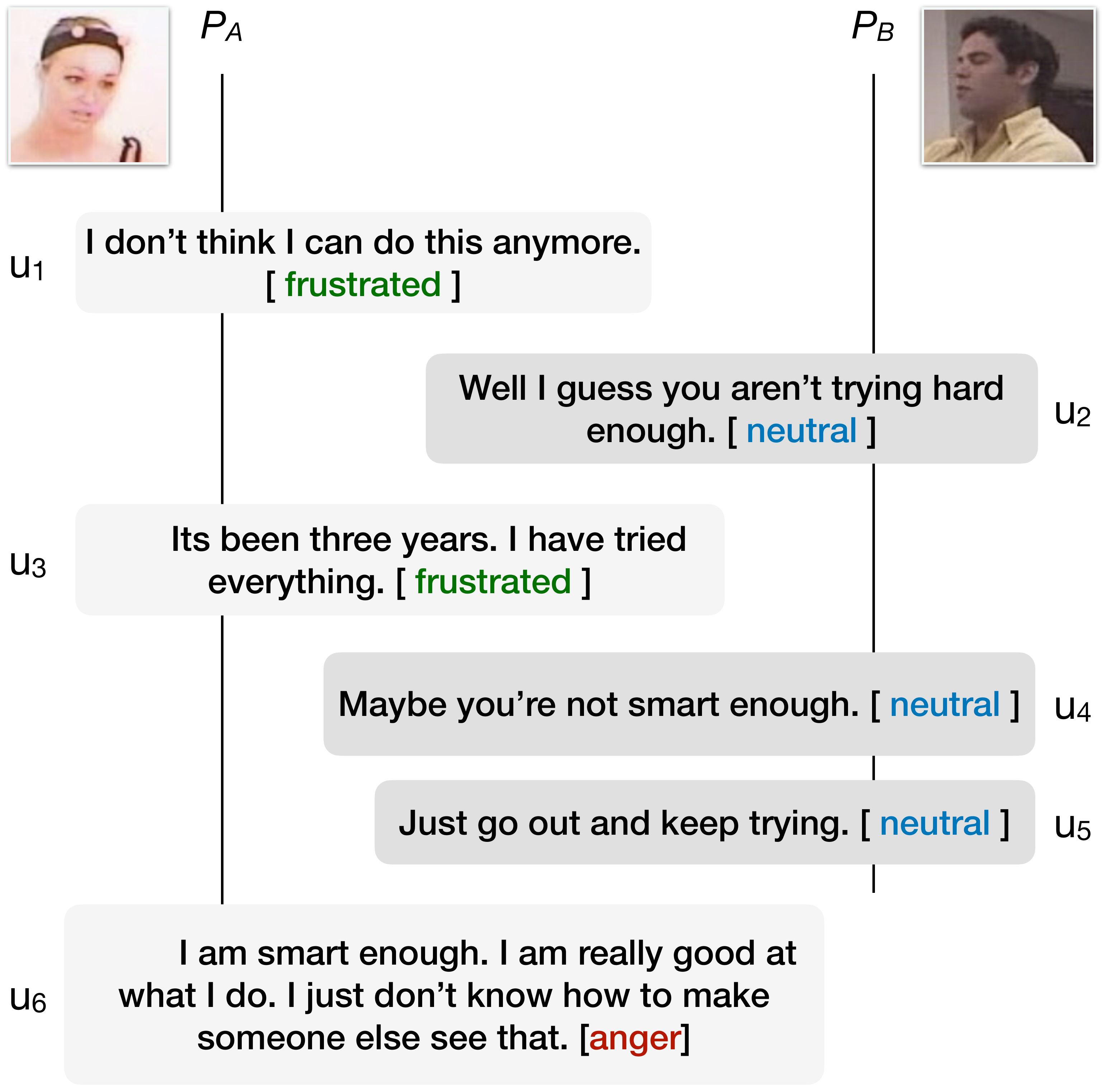} 
    \caption{An abridged dialogue from the IEMOCAP dataset~\citep{DBLP:journals/lre/BussoBLKMKCLN08}.}
    \label{fig:example}
\end{figure}

There have been very few works on inferring implicit sentiment~\citep{deng-wiebe-2014-sentiment} from context. This is crucial for achieving true sentiment understanding. Let us consider this sentence ``\emph{Oh no. The bill has been passed}''. As there are no explicit sentiment markers present in the isolated sentence -- \emph{``The bill has been passed''}, it would sound like a neutral sentence. Consequently, the sentiment behind `bill' is not expressed by any particular word. However, considering the sentence in the context -- \emph{``Oh no''}, which exhibits negative sentiment, it can be inferred that the opinion expressed on the `bill' is negative. The inferential logic that one requires to arrive at such conclusions is the understanding of sentiment flow in the context. In this particular example, the contextual sentiment of the sentence -- ``\textit{Oh no}'' flows to the next sentence and thus making it a negative opinionated sentence.
Tackling such tricky and fine-grained cases require bespoke modeling and datasets containing an ample quantity of such non-trivial samples. Further, commonsense knowledge can also aid in making such inferences. In the literature~\citep{poria2017context}, the use of LSTMs to model such sequential sentiment flow has been ineffectual. We think it would be fruitful to utilize logic rules, finite-state transducers, belief, and information propagation mechanisms to address this problems~\citep{deng2014joint,deng-wiebe-2014-sentiment}. We also note that contextual sentences may not always help. Hence, one can ponder using a gate or switch to learn and further infer when to count on contextual information.

In conversational sentiment-analysis, to determine the emotions and sentiments of an utterance at time $t$, the preceding utterances at time $<t$ can be considered as its context. However, computing this context representation can often be difficult due to complex sentiment dynamics.

Sentiments in conversations are deeply tied with emotional dynamics consisting of two important aspects: \textit{self-} and \textit{inter-personal dependencies}~\citep{morris2000emotions}. Self-dependency, also known as \textit{emotional inertia}, deals with the aspect of influence that speakers have on themselves during conversations~\citep{kuppens2010emotional}. On the other hand, inter-personal dependencies relate to the sentiment-aware influences that the counterparts induce into a speaker. Conversely, during the course of a dialogue, speakers also tend to mirror their counterparts to build rapport~\citep{navarretta2016mirroring}. This phenomenon is illustrated in Figure \ref{fig:example}. Here, $P_a$ is frustrated over her long term unemployment and seeks encouragement ($u_1, u_3$). $P_b$, however, is pre-occupied and replies sarcastically ($u_4$). This enrages $P_a$ to appropriate an angry response ($u_6$). In this dialogue, self-dependencies are evident in $P_b$, who does not deviate from his nonchalant behavior. $P_a$, however, gets sentimentally influenced by $P_b$. Modeling self- and inter-personal relationships and dependencies may also depend on the topic of the conversation as well as various other factors like argument structure, interlocutors’ personality, intents, viewpoints on the conversation, attitude towards each other, and so on. Hence, analyzing all these factors is key for a true self and inter-personal dependency modeling that can lead to enriched context understanding~\citep{DBLP:conf/naacl/HazarikaPZCMZ18}. 

The contextual information can come from both local and distant conversational history. As opposed to the local context, distant context often plays a smaller role in sentiment analysis of conversations. Distant contextual information is useful mostly in the scenarios when a speaker refers to earlier utterances spoken by any of the speakers in the conversational history.

\begin{figure*}[ht!] 
    \centering 
    \includegraphics[width=0.8\linewidth]{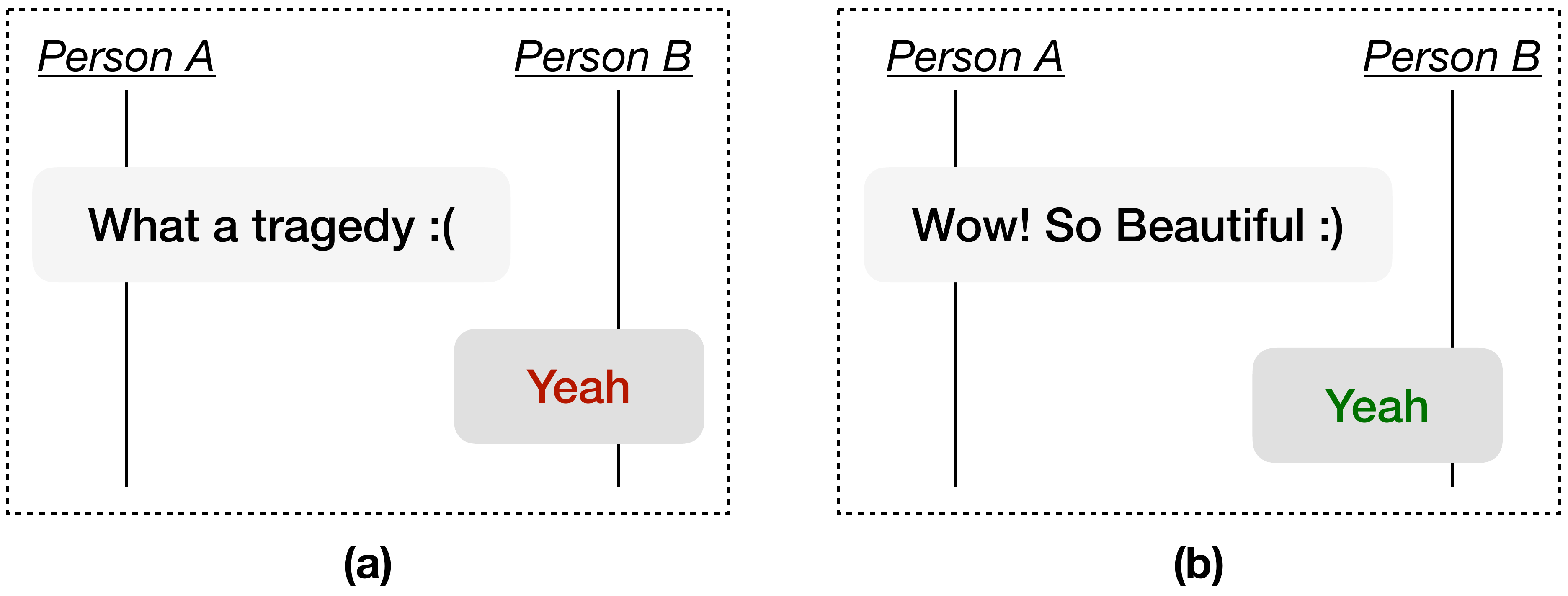}
    \caption{Role of context in sentiment analysis in conversation.}
    \label{fig:context_sentiment}
\end{figure*}

The usefulness of context is more prevalent in classifying short utterances, like {\it yeah}, {\it okay}, {\it no}, that can express different sentiments depending on the context and discourse of the dialogue. The examples in~\cref{fig:context_sentiment} explain this phenomenon. The sentiment expressed by the same utterance \textit{``Yeah"} in both these examples differ from each other and can only be inferred from the context. 

Leveraging such contextual clues is a difficult task. Memory networks, RNNs, and attention mechanisms have been used in previous works, e.g., HRLCE~\citep{DBLP:conf/semeval/HuangTZ19} or DialogueRNN~\citep{DBLP:conf/aaai/MajumderPHMGC19}, to grasp information from the context. However, these models fail to explain the situations where contextual information is needed. Hence, finding contextualized conversational utterance representations is an active area of research. 

\subsubsection{User, Cultural, and Situational Context}

Sentiment also depends on the user, cultural, and situational context. 

Individuals have subtle ways of expressing emotions and sentiments. For instance, some individuals are more sarcastic than others. For such cases, the usage of certain words would vary depending on if they are being sarcastic. Let's consider this example, $P_a:$ \textit{The order has been cancelled.}, $P_b:$ \textit{This is great!}. If $P_b$ is a sarcastic person, then his response would express negative emotion to the order being canceled through the word \textit{great}. On the other hand, $P_b$'s response, \textit{great}, could be taken literally if the canceled order is beneficial to $P_b$ (perhaps $P_b$ cannot afford the product he ordered). As necessary background
information is often missing from the conversations, speaker profiling based on preceding utterances often yields improved results. 

The underlying emotion of the same word can vary from one person to another. E.g., the word \emph{okay} can bear different sentiment intensity and polarity depending on the speaker's character. This incites the need to do user profiling for fine-grained sentiment analysis, which is a necessary task for e-commerce product review understanding. 

Understanding sentiment also requires cultural and situational awareness. A \textit{hot} and \textit{sunny} weather can be treated as a good weather in USA but certainly not in Singapore. Eating \textit{ham} could be accepted in one religion and prohibited by another.

A basic sentiment analysis system that only relies on distributed word representations and deep learning frameworks are susceptible to these examples if they do not encompass rudimentary contextual information.
\subsubsection{Role of Commonsense Knowledge in Sentiment Analysis}
\begin{figure}[ht!]
    \centering
    \includegraphics[width=.6\linewidth]{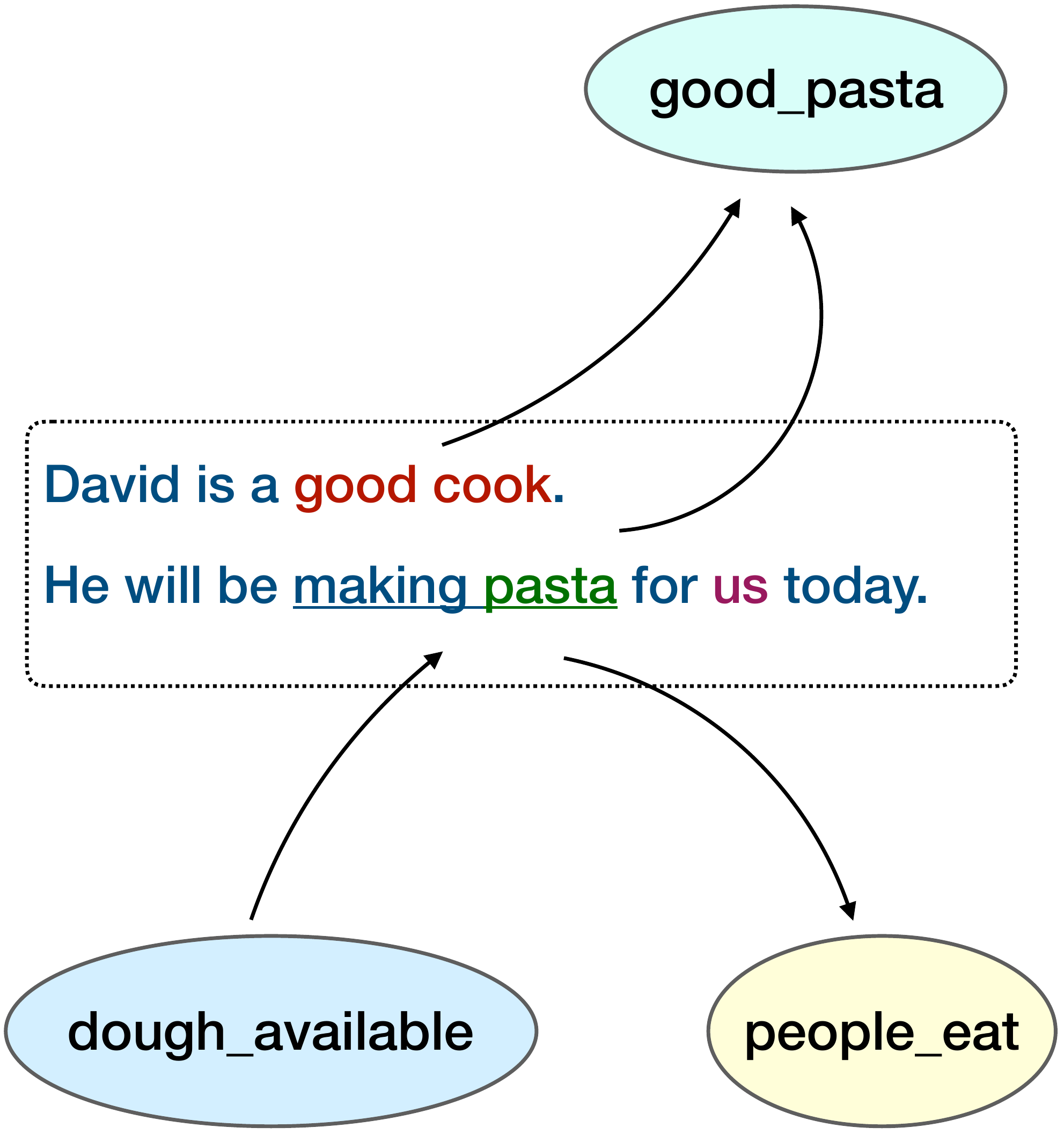}
    \caption{An illustration of commonsense reasoning and inference.}
    \label{fig:commonsense_inference}
\end{figure}

In layman's term, commonsense knowledge consists of facts that all human beings are expected to know. Due to this characteristic, humans tend to ignore expressing commonsense knowledge explicitly. As a result, word embeddings trained on the human-written text do not encode such trivial yet important knowledge that can potentially improve language understanding. The distillation of commonsense knowledge, thus, has become a new trend in modern NLP research. We show one such example in the \cref{fig:commonsense_inference} which illustrates the latent commonsense concepts that humans easily infer or discover given a situation. In particular, the present scenario informs that David is a good cook and will be making pasta for some people. Based on this information, commonsense can be employed to infer related events such as, dough for the pasta would be available, people would eat food (pasta), the pasta is expected to be good (David is good cook), etc. These inferences would enhance the text representation with many more concepts that can be utilized by neural systems in diverse downstream tasks.

\begin{figure*}[ht!] 
    \centering 
    \includegraphics[width=0.95\linewidth]{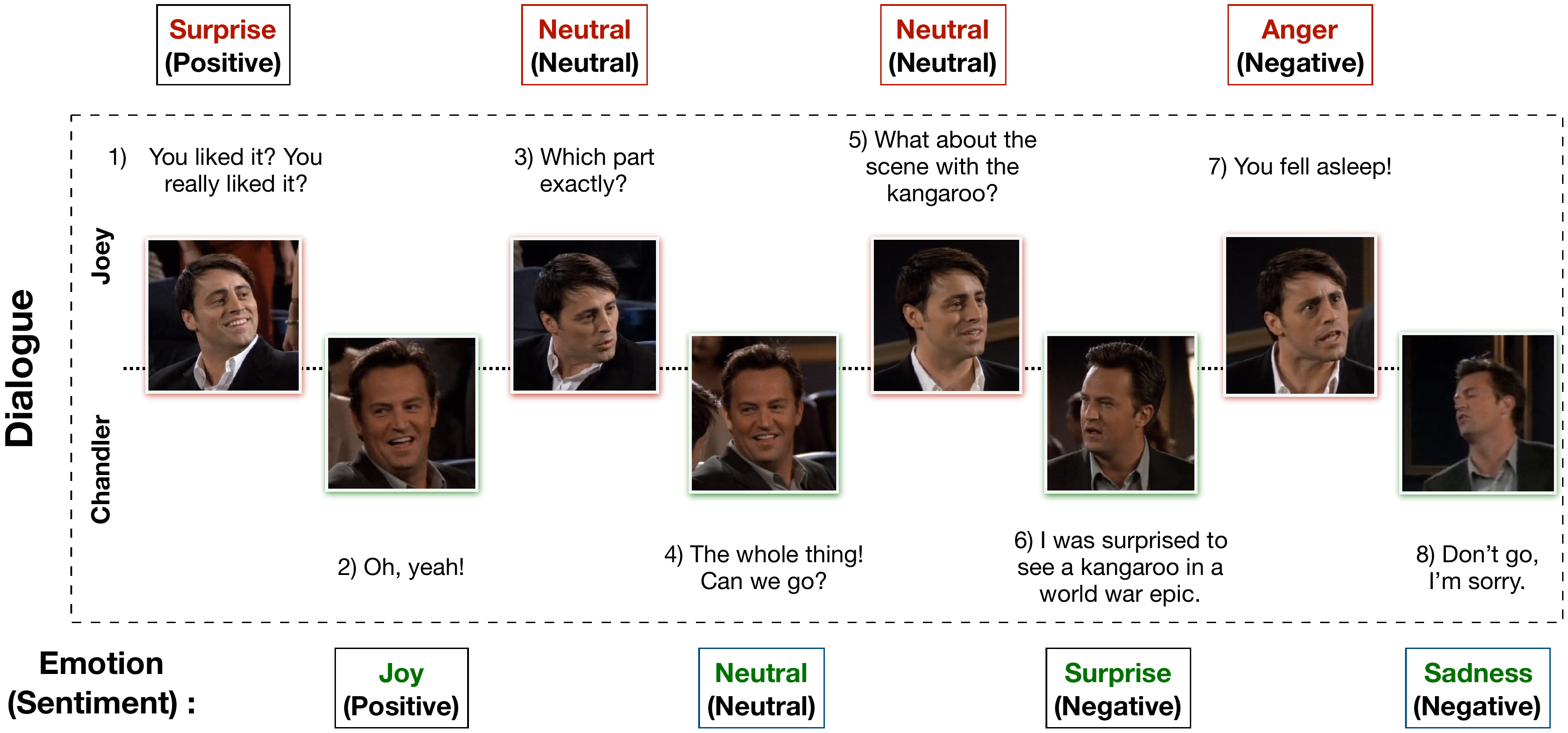}
    \caption{ 
Sentiment cause analysis.}
    \label{fig:exampleshift0}
\end{figure*}

In the context of sentiment analysis, utilizing commonsense for associating aspects with their sentiments can be highly beneficial for this task.
Commonsense knowledge graphs connect the aspects to various sentiment-bearing concepts via semantic links~\citep{DBLP:conf/aaai/MaPC18}. Additionally, semantic links between words can be utilized to mine associations between the opinion target and the opinion-bearing word.
What is the best way to grasp commonsense knowledge is still an open research question. 

Commonsense knowledge is also required to understand \emph{implicit} sentiment of the sentences that do not accommodate any explicit sentiment marker. E.g., the sentiment of the speaker in this sentence, \emph{``We have not seen the sun since last week''} is negative as not catching the sight of the sun for a long time is generally treated as a negative event in our society. A system not adhering to this commonsense knowledge would fail to detect the underlying sentiment of such sentences correctly. 

With the advent of commonsense modeling algorithms such as Comet~\citep{bosselut2019comet}, we think, there will be a new wave of research focusing on the role of commonsense knowledge in sentiment analysis in the near future.

\subsection{Sentiment Reasoning}

Apart from exploring the \textit{what}, we should also explore the \textit{who} and \textit{why}. Here, the \textit{who} detects the entity whose sentiment is being determined, whereas \textit{why} reveals the stimulus/reason for the sentiment.

\subsubsection{Who? The Opinion Holder} 

While analyzing opinionated text, it is often important to know the opinion holder. In most cases, the opinion holder is the person who spoke/wrote the sentence. Yet, there can be situations where the opinion holder is an entity (or entities) mentioned in the text~\citep{mohammad2017challenges}. Consider the following two lines of opinionated text:
\begin{enumerate}[a.]
    \item \textit{The movie was too slow and boring.}
    \item \textit{Stella found the movie to be slow and boring.}
\end{enumerate}
In both the sentences above, the sentiment attached to the movie is negative. However, the opinion holder for the first sentence is the speaker while in the second sentence it is Stella. The task could be further complex with the need to map varied usage of the same entity term (e.g. Jonathan, John) or the use of pronouns (he, she, they)~\citep{liu2012sentiment}.

Many works have studied the task of opinion-holder identification -- a subtask of opinion extraction (opinion holder, opinion phrase, and opinion target identification). These works include approaches that use named-entity recognition~\citep{kim2004determining}, parsing and ranking candidates~\citep{kim2006extracting}, semantic role labeling~\citep{DBLP:conf/conll/WiegandR15}, structured prediction using CRFs~\citep{DBLP:conf/emnlp/ChoiBC06}, multi-tasking~\citep{DBLP:conf/acl/YangC13}, amongst others. The MPQA corpus~\citep{DBLP:conf/naacl/DengW15} provided supervised annotations for this task. However, for deep learning approaches, this topic has been understudied~\citep{DBLP:conf/naacl/ZhangLF19,DBLP:conf/bigdataconf/QuanZH19}.

\subsubsection{Why? The Sentiment Stimulus}
The majority of the sentiment analysis research works to date are about classifying contents into positive, negative, and neutral and till date, only a little attention has been paid to sentiment cause identification. 
Future research in sentiment analysis should focus on what drives a person to express positive or negative sentiment on a topic or aspect.

To reason about a particular sentiment of an opinion-holder, it is important to understand the target of the sentiment~\citep{deng-wiebe-2014-sentiment}, and whether there are implications of holding such sentiment. For instance, when stating ``\textit{I am sorry that John Doe went to prison.}'', understanding the target of the sentiment in the phrase "\textit{John Doe goes to prison}", and knowing that ``\textit{go to prison}'' has negative implications on the target, implies positive sentiment toward John Doe.\footnote{Example provided by Jan Wiebe (2016), personal communication.} Moreover, it is important to understand what caused the sentiment. Although in this example, it is straightforward to conclude that ``\textit{go to prison}'' is the reason of the expressed negative expression. One can also infer new knowledge from this simplified reasoning --- ``\textit{go to prison}'' is a negative event. This sentiment knowledge discovery can further help to enrich phrase-level sentiment lexicons.

The ability to reason is necessary for any explainable AI system. In the context of sentiment analysis, it is often desired to understand the cause of an expressed sentiment by the speaker. E.g., consider a review on a smartphone, ``\textit{I hate the touchscreen as it freezes after 2-3 touches}''. While it is critical to detect the negative sentiment expressed on \emph{touchscreen}, digging into the detail that causes this sentiment is also of prime importance~\citep{liu2012sentiment}, which in this case is implied by the phrase ``\textit{freezes after 2-3 touches}''. To date, there is not much work exploring this aspect of the sentiment analysis research. \citet{li2017reflections} discuss two possible reasons that give rise to opinions. Firstly, an opinion-holder might have an emotional bias towards the entity/topic in question. Secondly, the sentiment could be borne out of mental (dis)satisfaction towards a goal achievement. 

Grasping the cause of sentiment is also very important in dialogue systems. As an example, we can refer to Figure~\ref{fig:exampleshift0}, where Joey expresses \emph{anger} once he ascertains Chandler's \emph{deception} in the previous utterance.

It is hard to define a taxonomy or tagset for the reasoning of both emotions and sentiments. At present, there is no available dataset that contains such rich annotations. Building such a dataset would enable future dialogue systems to frame meaningful argumentation logic and discourse structure, taking one step closer to human-like conversation.


\subsection{Domain Adaptation}

Most of the state-of-the-art sentiment analysis models enjoy the privilege of having in-domain training datasets. However, this is not a viable scenario, as curating large amounts of training data for every domain is impractical. 
Domain adaptation in sentiment analysis solves this problem by learning the characteristics of the unseen domain. 
Sentiment classification, in fact, is known to be sensitive towards domains as mode of expressing opinions across domains vary. Also, valence of affective words may vary based on different domains~\citep{liu2012sentiment}.

Diverse approaches have been proposed for cross-domain sentiment analysis. One line of work models domain-dependent word embeddings~\citep{DBLP:conf/acl/SarmaLS18, DBLP:conf/acl/LamSBF18,k-sarma-etal-2019-shallow} or domain-specific sentiment lexicons~\citep{DBLP:conf/emnlp/HamiltonCLJ16}, while others attempt to learn representations based on either co-occurrences of domain-specific with domain-independent terms (pivots)~\citep{DBLP:conf/acl/BlitzerDP07, DBLP:conf/www/PanNSYC10, DBLP:conf/naacl/ZiserR18, DBLP:conf/acl/BhattacharyyaDS18} or shared representations using deep networks~\citep{glorot2011domain}.  

One of the major breakthroughs in domain adaptation research employs adversarial learning that trains to fool a domain discriminator by learning domain-invariant representations~\citep{JMLR:v17:15-239}. In this work, the authors utilize bag of words as the input features to the network. Incorporating bag of words limits the network to access any external knowledge about the unseen words of the target domain. Hence, the performance improvement can be completely attributed to the efficacy of the adversarial network. However, in recent works, researchers tend to utilize distributed word representations such as Glove, BERT. These representations, aka word embeddings, are usually trained on huge open-domain corpora and consequently contain domain invariant information. Future research should explain whether the gain in domain adaptation performance comes from these word embeddings or the core network architecture. 

In summary, the works in domain adaptation lean towards outshining the state of the art on benchmark datasets. What remains to be seen is the interpretability of these methods. Although some works claim to learn the domain-dependent sentiment orientation of the words during domain invariant training, there is barely any well-defined analysis to validate such claims. 

\subsubsection{Use of External Knowledge}
\begin{figure}[t]
    \centering
    \includegraphics[width=\linewidth]{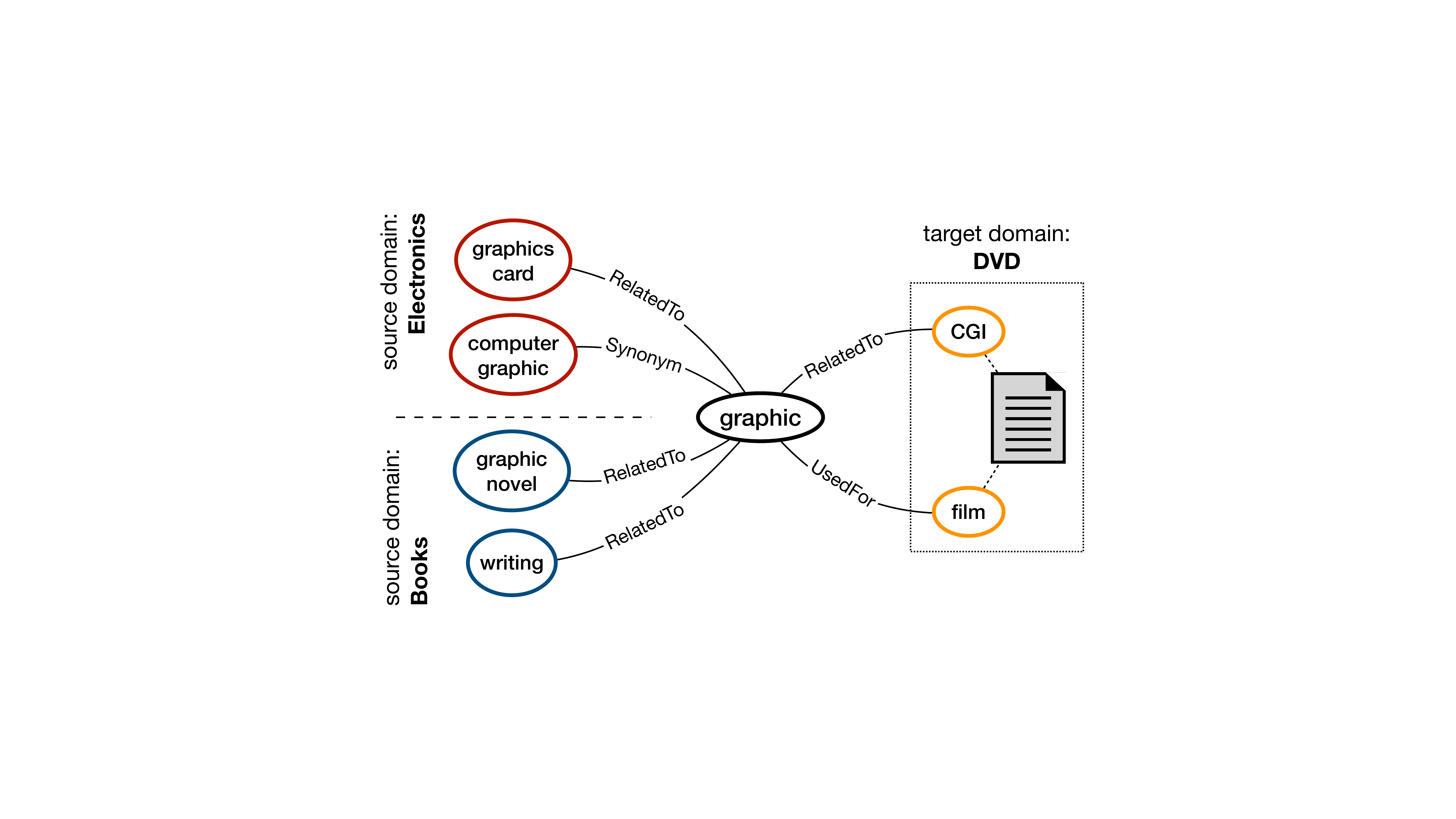}
    \caption{Domain-general term \textit{graphic} bridges the semantic knowledge between domain specific terms in Electronics, Books and DVD.}
    \label{fig:case_study}
\end{figure}

The key idea that most of the existing works encapsulate is to learn domain-invariant shared representations as a means to domain adaptation. While global or contextual word embeddings have shown their efficacy in modeling domain-invariant and specific representations, it might be a good idea to couple these embeddings with multi-relational external knowledge graphs for domain adaptation. Multi-relation knowledge graphs represent semantic relations between concepts. Hence, they can contain complementary information over the word embeddings, such as Glove, since these embeddings are not trained on explicit semantic relations. Semantic knowledge graphs can establish relationships between domain-specific concepts of several domains using domain-general concepts -- providing vital information that can be exploited for domain adaptation. One such example is presented in \cref{fig:case_study}. Researchers are encouraged to read these early works~\citep{DBLP:conf/acl/JotyAI18,DBLP:journals/tkde/XiangCHY10} on exploiting external knowledge for domain adaptation.

\subsubsection{Scaling Up to Many Domains}
Most of the present works in this area use the setup of a \textit{source} and \textit{target} domain pair for training. Although appealing, this setup requires retraining as and when the target domain changes. The recent literature in domain adaptation goes beyond single-source-target~\citep{DBLP:conf/nips/ZhaoZWMCG18} to multi-source and multi-target~\citep{DBLP:journals/tip/GholamiSRBP20,DBLP:journals/corr/abs-1809-00852} training. However, in sentiment analysis, these setups have not been fully explored and deserve more attention~\citep{DBLP:conf/acl/WuH16}.

\subsection{Multilingual Sentiment Analysis} \label{sec:multilingual}

The majority of sentiment analysis research has been conducted on English datasets. However, the advent of social media platforms has made multilingual content available via platforms such as Facebook and Twitter. Consequently, there is a recent surge in works with diverse languages~\citep{DBLP:journals/cogcom/DashtipourPHCHG16a}. The NLP community, in general, is now also vocal to promote research on languages other than English.\footnote{Because of a now widely known statement made by Professor Emily M.Bender on Twitter, 
we now use the term \emph{\#BenderRule} to require that the language addressed by research projects by explicitly stated, even when that language is English \url{https://bit.ly/3aIqS0C}}

In the context of sentiment analysis, despite the recent surge in multilingual sentiment analysis, several directions need more traction:

\subsubsection{Language-Specific Lexicons}
Today's rule-based sentiment analysis system, such as Vader, works great for the English language, thanks to the availability of resources like sentiment lexicons. For other languages such as Hindi, French, Arabic, not many well-curated lexicons are available.

\subsubsection{Sentiment Analysis of Code-Mixed Data}
In many cultures, people on social media post content that are a mix of multiple languages~\citep{lal2019mixing, Guptha2020, gamback-das-2016-comparing}. For example, ``\textbf{\textit{Itna izzat diye aapne mujhe
!!!}} Tears of joy. :’( :’(", in this sentence, the bold text is in Hindi with roman orthography, and the rest is in English. Code-mixing poses a significant challenge to the rule- and deep learning-based methods. A possible future work to combat this challenge would be to develop language models on code-mixed data. How and where to mix languages is a person's own choice, one of the main hardships. Another critical challenge associated with this task is identifying the deep compositional semantic that lies in the code mixed data. Unfortunately, only a little research has been carried out on this topic~\citep{lal2019mixing,DBLP:conf/coling/JoshiPSV16}.

\subsubsection{Machine Translation as a Solution to Multilingual Sentiment Analysis}

Can machine translation be used as a solution to multilingual or cross-lingual sentiment analysis? Recently, several papers~\cite{saadany2020great,balamurali2013lost} have attempted this problem. \citet{saadany2020great} claim that the associated sentiment is not preserved for contronyms, negations, diacritic, and idiomatic expressions when translated from Arabic to English. One solution to tackle this problem, as proposed by \citet{saadany2020great}, is integrating sentiment information in the encoding stage in a machine translation system. However, this research has yet to witness much research attention.

\subsection{Sarcasm Analysis}

The study of sarcasm analysis is highly integral to the development of sentiment analysis due to its prevalence in opinionated text~\citep{maynard2014cares,DBLP:journals/expert/MajumderPPCCGC19}. Detecting sarcasm is highly challenging due to the figurative nature of text, which is accompanied by nuances and implicit meanings~\citep{jorgensen1984test}. Over recent years, this research field has established itself as an important problem in NLP with many works proposing different solutions to address this task~\citep{joshi2017automatic}. Broadly, the main contributions have emerged from the speech and text communities. In speech, existing works leverage different signals such as prosodic cues~\citep{bryant2010prosodic,woodland2011context}, acoustic features including low-level descriptors, and spectral features~\citep{cheang2008sound}. Whereas in textual systems, traditional approaches consider rule-based~\citep{khattri2015your} or statistical patterns~\citep{gonzalez2011identifying}, stylistic patterns~\citep{tsur2010icwsm}, incongruity~\citep{joshi2015harnessing,DBLP:conf/acl/SuTHL18}, situational disparity~\citep{riloff2013sarcasm},  and hashtags~\citep{maynard2014cares}. While stylistic patterns, incongruity, and valence shifters are some of the ways humans use to express sarcasm, it is also highly contextual. In addition, sarcasm also depends on a person's personality, intellect, and the ability to reason over commonsense. In the literature, these aspects of sarcasm remain under-explored.
\begin{figure}[t]
	\includegraphics[width=\linewidth]{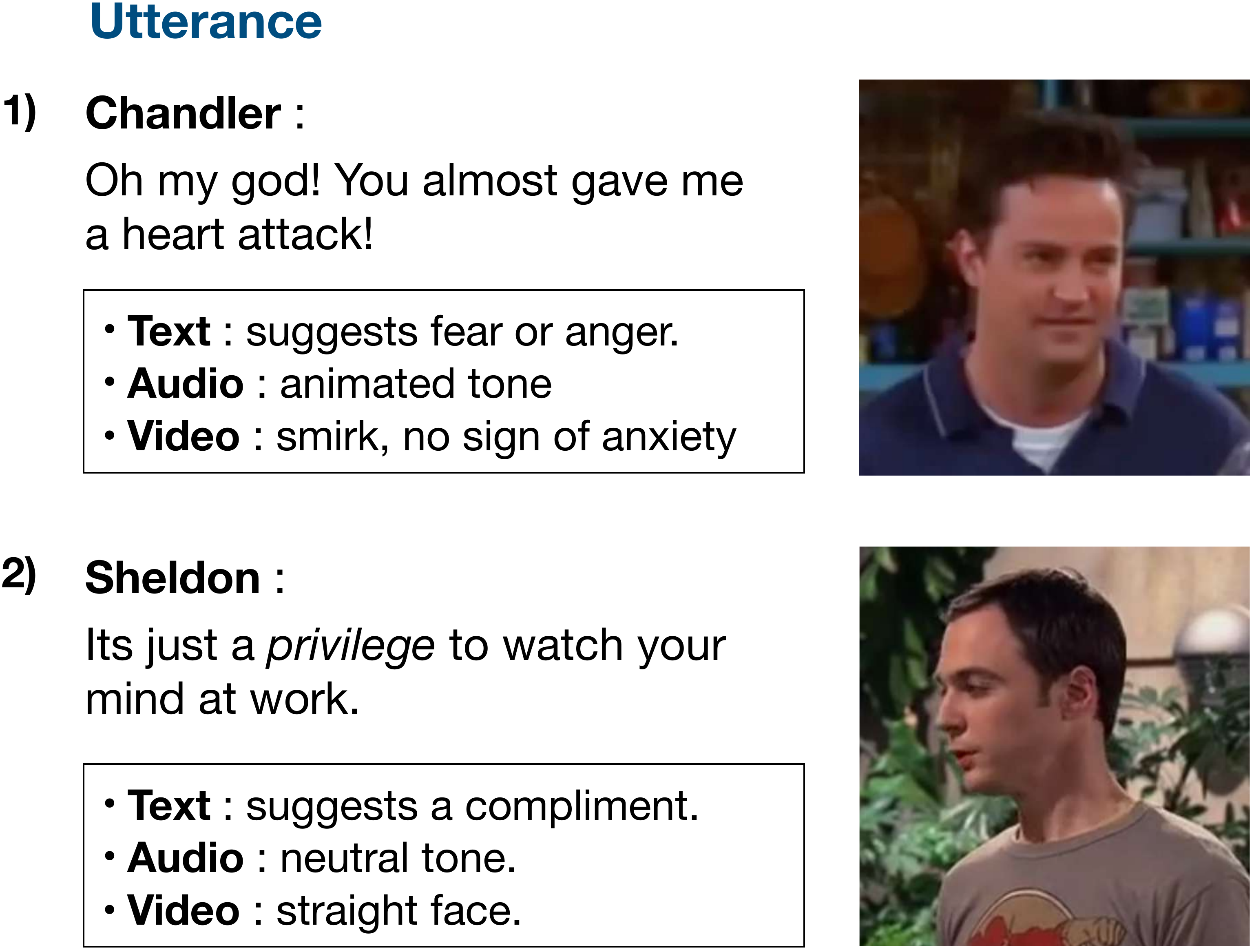}
	\caption{Incongruent modalities in sarcasm present in the MUStARD dataset~\citep{DBLP:conf/acl/CastroHPZMP19}.}
	\label{fig:multimodal_examples2}
\end{figure}
\subsubsection{Leveraging Context in Sarcasm Detection}
Although the research for sarcasm analysis has primarily dealt with analyzing the sentence at hand, recent trends have started to acquire contextual understanding by looking beyond the text. Similar to sentiment analysis (~\cref{sec:multimodal_sa,sec:contextual_sa}), sarcasm detection can benefit with contextual cues provided by conversation histories, author tendencies, and multimodality.

\paragraph*{\textbf{User Profiling and Conversational Context}}
Two types of contextual information have been explored for providing additional cues to detect sarcasm: \textit{authorial context} and \textit{conversational context}. Leveraging authorial context delves with analyzing the author's sarcastic tendencies (user profiling) by looking at their historical and meta data~\citep{bamman2015contextualized,hazarika2018cascade}. Similarly, the conversational context uses the additional information acquired from surrounding utterances to determine whether a sentence is sarcastic~\citep{DBLP:journals/coling/GhoshFM18}. It is often found that sarcasm is apparent only when put into context over what was mentioned earlier. For example, when tasked to identify whether the sentence ``\textit{He sure played very well}" is sarcastic, it is imperative to look at prior statements in the conversation to reveal facts (``\textit{The team lost yesterday}").


\paragraph*{\textbf{Multimodal Context}}
We also identify \textit{multimodal signals} to be important for sarcasm detection. Sarcasm is often expressed without linguistic markers, and instead, by using verbal and non-verbal cues. Change of tone, overemphasis on words, straight face, etc. are some such cues that indicate sarcasm. There have been very few works that adopt multimodal strategies to determine sarcasm~\citep{schifanella2016detecting}. \citet{DBLP:conf/acl/CastroHPZMP19} recently released a multimodal sarcasm detection dataset, MUStARD, that takes conversational context into account. 
\cref{fig:multimodal_examples2} presents two cases from this dataset, where sarcasm is expressed through the incongruity between modalities. In the first case, the language modality indicates fear or anger. In contrast, the facial modality lacks any visible sign of anxiety that would agree with the textual modality. In the second case, the text is indicative of a compliment, but the vocal tonality and facial expressions show indifference. In both cases, the incongruity between modalities acts as a strong indicator of sarcasm. While useful, MUStARD contains only 500 odd instances, posing a significant challenge to training deep networks on this dataset. 

\subsubsection{Annotation Challenges: Intended vs. Perceived Sarcasm}

Sarcasm is a highly subjective tool and poses significant challenges in curating annotations for supervised datasets. This difficulty is particularly evident in \textit{perceived sarcasm}, where human annotators are employed to label text as sarcastic or not. Sarcasm recognition is known to be a difficult task for humans due to its reliance on pragmatic factors such as common ground~\citep{DBLP:books/cu/C1996}. This difficulty is also observed through the low annotator agreements across the datasets curated for perceived sarcasm~\citep{DBLP:conf/acl/Gonzalez-IbanezMW11,DBLP:conf/acl/CastroHPZMP19}. To combat such perceptual subjectivity, recent emotion analysis approaches utilize \textit{perceptual uncertainty} in their modeling~\citep{zhang2018dynamic, gui2017curriculum, han2017hard}.

In our experience of curating a multimodal sarcasm detection dataset~\citep{DBLP:conf/acl/CastroHPZMP19}, we observed poor annotation quality, which occurred mainly due to the hardships associated with this task.
\citet{DBLP:conf/naacl/HovyBVH13} noticed that people undertaking such tasks remotely online are often guilty of \textit{spamming}, or providing careless or random responses.

One solution to this problem is to rely on self annotated data collection. While convenient, obtaining labeled data from hashtags has been found to introduce both noise (incorrectly-labeled examples) and bias  (only certain forms of sarcasm are likely to be tagged~\citep{davidov2010semi}, and predominantly by certain types of Twitter users~\citep{bamman2015contextualized}). 

Recently, \citet{DBLP:journals/corr/abs-1911-03123} presented the iSarcasm dataset, which provides labels by the original writers for the sarcastic posts. This kind of annotation is promising as it circumvents the issues mentioned above while capturing the intended sarcasm. To address the issues stemmed from annotating perceived sarcasm, Best-Worst Scaling (MaxDiff/BWS)~\citep{DBLP:conf/naacl/KiritchenkoM16} could be employed to alleviate the effect of subjectivity in annotations. BWS attempts to alleviate the ambiguity in annotations by asking annotators to compare rather than indicate. Thus, rather than identifying the presence of sarcasm or its intensity, BWS would ask annotators to choose the most sarcastic (best) and least sarcastic (worst) sentences from candidate 4-tuples -- leading to easier and better decision making. \citet{DBLP:conf/naacl/KiritchenkoM16} shows that such comparisons can be converted into a ranked list of the items based on the property of interest, which in this case is perceived sarcasm.

\subsubsection{Target Identification in Sarcastic Text} 

Identifying the target of ridicule within a sarcastic text -- a new concept recently introduced by~\citet{DBLP:conf/lrec/JoshiGBC18} -- has important applications. It can help chat-based systems better understand user frustration and help ABSA tasks assign the sarcastic intent with the correct target in general. Though similar, there are differences from the vanilla aspect extraction task (\cref{sec:absa}) as the text might contain multiple aspects/entities with only a subset being a sarcastic target~\citep{DBLP:conf/emnlp/PatroBM19}. When expressing sarcasm, people tend not to use the target of ridicule explicitly, which makes this task immensely challenging to combat.

\subsubsection{Style Transfer between Sarcastic and Literal Meaning}

\paragraph*{\textbf{Figurative to Literal Meaning Conversion}}
Converting a sentence from its figurative meaning to its honest and literal form is an exciting application.  It involves taking a sarcastic sentence such as \textit{``I loved sweating under the sun the whole day"} to \textit{``I hated sweating under the sun the whole day"}.  It has the potential to aid opinion mining, sentiment analysis, and summarization systems. These systems are often trained to analyze the literal semantics, and such a conversion would allow for accurate processing. Present approaches include converting a full sentence using monolingual machine translation techniques~\citep{DBLP:conf/acl/PeledR17}, and also word-level analysis, where target words are disambiguated into their sarcastic or literal meaning~\citep{DBLP:conf/emnlp/GhoshGM15}.  This application could also help in 1) performing data augmentation and 2) generating adversarial examples as both the forms (sarcastic and literal) convey the same meaning but with different lexical forms.

\paragraph*{\textbf{Generating Sarcasm from Non-Figurative Sentences}}

The ability to generate sarcastic sentences is an important yardstick in the development of NLG. The goal of building socially-relevant and engaging, interactive systems demand such creativity. Sarcastic content generation can also benefit content/media generation that finds applications in fields like advertisements. \citet{DBLP:conf/emnlp/MishraTS19} recently proposed a modular approach to generate sarcastic text from negative sentiment-aware scenarios. End-to-end counterparts to this approach have not been well studied yet. Also, most of the works here rely on a particular type of sarcasm -- one which involves incongruities within the sentence. The generation of other flavors of sarcasm (as mentioned before) has not been yet studied. Recently,~\citet{chakrabarty-etal-2020-r} proposed a retrieval-based method that focuses on some of the points raised above, indicating the promise of research in this direction.

\subsubsection{Sentiment and Creative Language}

While the above discussion is focused on sarcasm analysis, other forms of creative language tools, such as irony, humor, etc. also have co-dependence with sentiment analysis. \textit{Irony} is more generic than sarcasm as it used to mean the opposite of what is being said, whereas sarcasm has an intent to criticize. Previous works have utilized sentiment information for detecting irony~\cite{DBLP:conf/ibpria/FariasBR15}. However, compared to sarcasm, research utilizing sentiment-irony relationships has been scanty. \textit{Humor} is another tool that is often used in human language. Identifying humor also has deep ties with sentiment analysis. Multiple works mention that detecting humor often relies on mining the sentimental relations between the \textit{setup} and the corresponding \textit{punchline}~\cite{DBLP:conf/acl/LiuZS18,DBLP:conf/emnlp/HasanRZZTMH19,DBLP:journals/ipm/ZhangZCR19}. Recent works also reveal that architectures developed for sentiment analysis perform similarly for tasks related to humor prediction~\cite{DBLP:conf/mm/HazarikaZP20}. Looking at the reverse, some works demonstrate that knowing humor, sarcasm, etc. aids in making sentiment analysis systems more robust~\cite{badlani-etal-2019-ensemble}.
The above points indicate the relationships between sentiment analysis with creative language, thus opening whole new doors for research~\cite{DBLP:journals/expert/MajumderPPCCGC19}.


\subsection{Sentiment-Aware Natural Language Generation (NLG)}

Language generation is considered one of the major components of the field of NLP. Historically, the focus of statistical language models has been to create syntactically coherent text using architectures such as n-grams models~\citep{stolcke2002srilm} or auto-regressive recurrent architectures~\citep{bengio2003neural, mikolov2010recurrent, sundermeyer2012lstm}. These generative models have important applications in areas including representation learning, dialogue systems, amongst others. However, present-day models are not trained to produce affective content that can emulate human communication. Such abilities are desirable in many applications such as comment/review generation~\citep{DBLP:conf/eacl/ZhouLWDHX17}, and emotional chatbots~\citep{zhou2018emotional, DBLP:journals/inffus/MaNXC20}.

\begin{figure*}[t]
    \centering
    \begin{subfigure}{0.49\textwidth}
     \includegraphics[width=\linewidth]{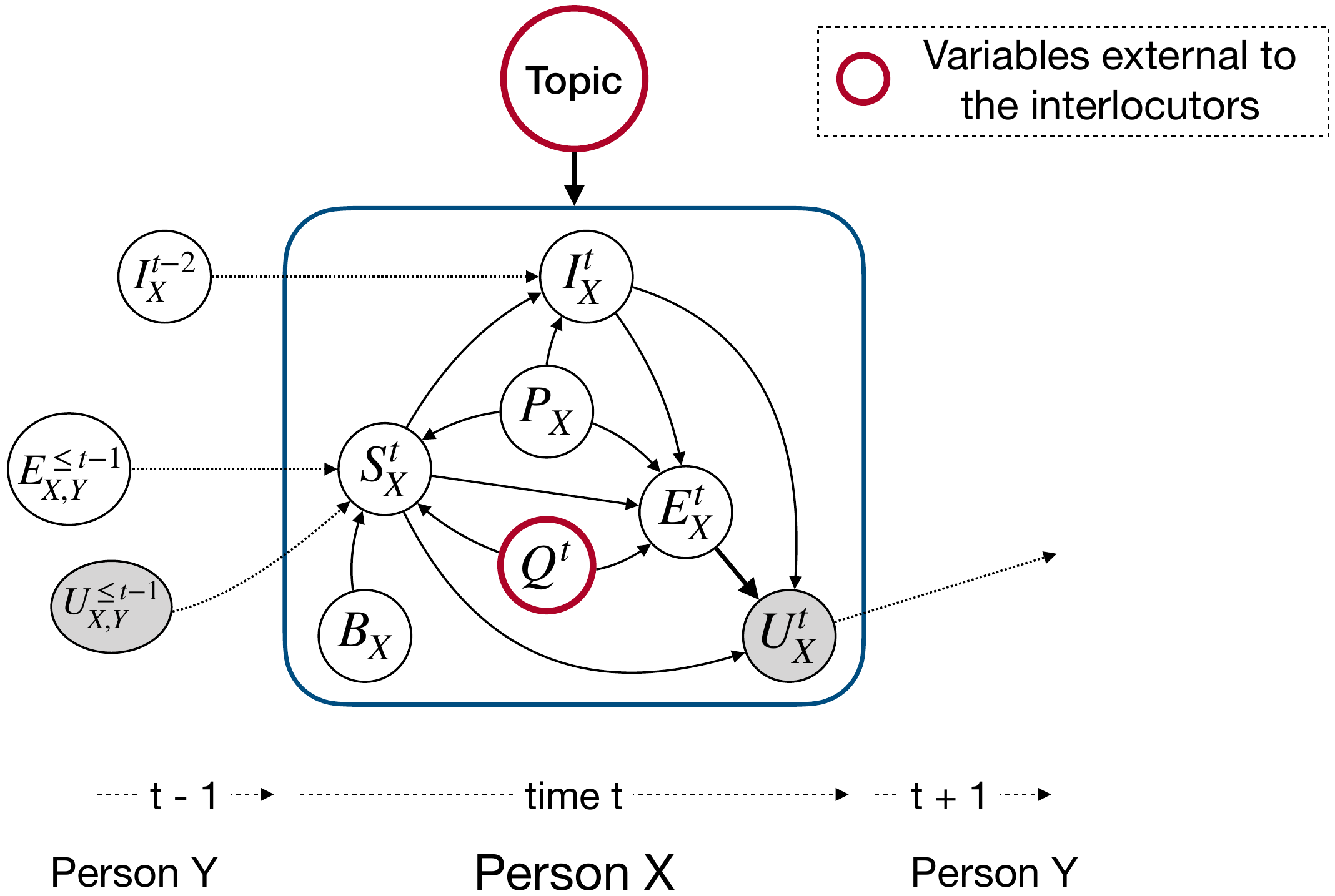}
      \label{fig:controlling_vars1}
      \caption{}
     \end{subfigure}
     \begin{subfigure}{0.49\textwidth}
     \includegraphics[width=\linewidth]{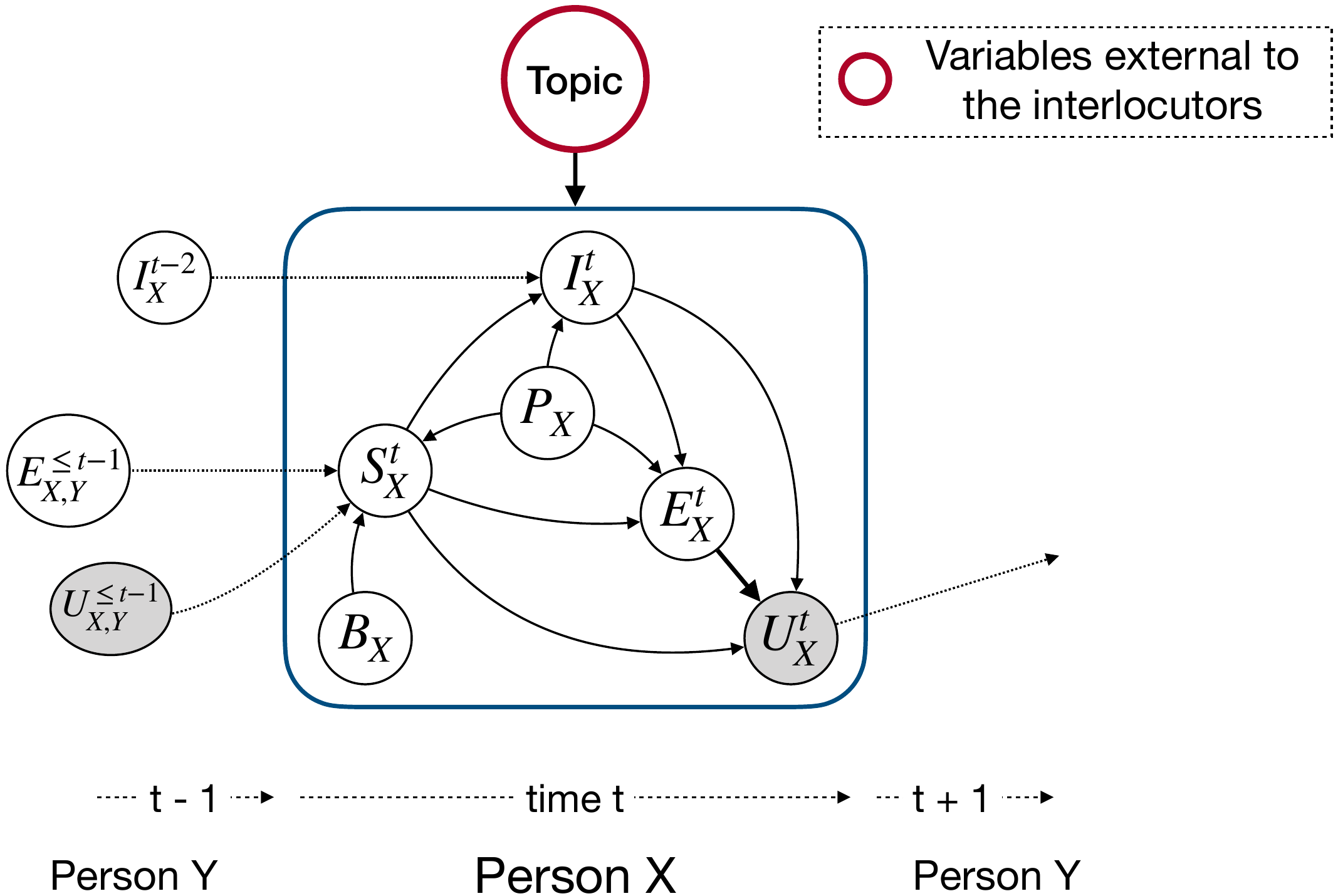}
      \label{fig:controlling_vars2}
      \caption{}
     \end{subfigure}
     \caption{Our proposed conceptual framework for conditional generative conversation modeling. A dyadic conversation between person X and Y are governed by interactions between several latent factors, such as, intent, emotion. (a) The complete emotion-aware conditional generative model that accounts for the variables both internal and external to the interlocutors. (b) A simplified version where the presence of external and sensory inputs are ignored.}
     \label{fig:controlling_vars}
\end{figure*}


Early efforts in this direction included works that either focused on related topics such as personality-conditioned text generation~\citep{mairesse2007personage} or pattern-based approaches for the generation of emotional sentences~\citep{keshtkar2011pattern}. These works were significantly pipe-lined with specific modules for sentence structure and content planning, followed by surface realization. Such sequential modules allowed constraints to be defined based on personality/emotional traits, which were mapped to sentential parameters that include sentence length, vocabulary usage, or part-of-speech (POS) dependencies. Needless to say, such efforts, though well-defined, are not scalable to general scenarios and cross-domain settings.

\subsubsection{Conditional Generative Models}

We, human beings, count on several variables such as emotion, sentiment, prior assumptions, intent, or personality to participate in dialogues and monologues. In other words, these variables control the language that we generate. Hence, it is outrageous to claim that a vanilla seq2seq framework can generate \emph{near perfect} natural language. Recently, conditional generative models have been developed to address this task. Conditioning on attributes, such as, sentiment can be approached in several ways. One way is by learning \textit{disentangled representations}, where the key idea is to separate the textual content from high-level attributes, such as, sentiment and tense in the hidden latent code. 
Present approaches utilize generative models such as VAEs~\citep{DBLP:conf/icml/HuYLSX17}, GANs~\citep{DBLP:conf/ijcai/Wang018} or Seq2Seq models~\citep{radford2017learning}. Learning disentangled representations is presently an open area of research. Enforcing independence of factors in the latent representation and presenting quantitative metrics to evaluate the factored hidden code are some of the challenges associated with these models. 

An alternate method is to pose the problem as an \textit{attribute-to-text} translation task~\citep{DBLP:conf/eacl/ZhouLWDHX17,DBLP:conf/inlg/ZangW17}. In this setup, desired attributes are encoded into hidden states which condition upon a decoder tasked to generate the desired text. The attributes could include user's preferences (including historical text), descriptive phrases (e.g. product description for reviews), and sentiment. Similar to general translation tasks, this approach demands parallel data and raises generalization challenges, such as, cross-domain generalization. Moreover, the attributes might not be available in the desired formats. As mentioned, attributes might be embedded in conversational histories, which would require sophisticated NLU capabilities similar to the ones used in task-oriented dialogue bots. They might also be in the form of structured data, such as Wikipedia tables or knowledge graphs, tasked to be translated into textual descriptions, i.e., data-to-text -- an open area of research~\citep{DBLP:conf/acl/MishraLSJK19}.

\paragraph{\textbf{Our conceptual conditional generative model}}

In \cref{fig:controlling_vars}, we illustrate a dialogue-generation mechanism that leverages these key variables. In this illustration, $P$ represents the personality of the speaker; $S$ represents the speaker-state; $I$ denotes the intent of the speaker; $E$ refers to the speaker's emotional-aware state, and $U$ refers to the observed utterances. The definitions of these variables are as given below:

\textbf{Topic:} Topic is a key element that governs and drives a conversation. Without knowing the topical information, dialogue understanding can be incomplete and vague.

\textbf{Personality ($P$):} As per the standard definition of personality, this variable controls the basic behavior of a person under varied circumstances. Personality can also signify different dimensions, such as, values, needs, goals, agency, and more, according to the theory of appraisals~\citep{ellsworth2003appraisal} in affective computing.

\textbf{Observed conversational history ($U$):} The observed utterances in the conversational history are represented as $U$. $U$ provides contextual information and play a critical role in constructing other states such as $S$ and $I$.

\textbf{Background knowledge ($B$):} Background knowledge represents the prior assumptions, pre-existing inter-speaker relations, speaker's knowledge and opinion about the topic and any other background or external information that are not explicitly present in the conversational history. Such knowledge usually evolves over time depending on how the speaker experiences the environment and interacts with it.

\textbf{Speaker-state ($S$):} Speaker-state can be defined as the latent memory of a speaker that evolves over the turns in the conversation. This memory contains the information obtained through the process of cognition and thinking. For example, let us consider this conversation:

A (excited and happy): \textit{You know I am getting married!}\\
\indent B (excited and happy): \textit{Wow! that's great news. Who is that lucky person? When is the ceremony?}

In this conversation, person B listens to person A (refer to variable $U$ in \cref{fig:controlling_vars}) and applies cognition and thinking to construct a latent memory which we call as $S$. $S$ can also rely on the background knowledge $B$.

\textbf{Intent ($I$):} Intent defines the goal that the speaker wants to achieve in the conversation. Intent can be triggered after cognition and thinking i.e., $S$.

In the above example, it is obvious that the intent of person B (i.e., intention to know who is person A marrying and the date of the ceremony) is governed by his/her cognition after hearing the statement by person A. Intent can heavily rely on the personality of the speaker.

\textbf{External and sensory inputs ($Q$):} In the process of a conversation, certain sensory or other external events can directly initiate cognition and affect. We call these inputs as $Q$. These inputs can often be non-verbal cues. Affective reactions to these sensory inputs can occur with or without any complex cognitive modeling. When the stimulus is sudden and unexpected, the affective reaction can occur before evaluating and appraising the situation through cognitive modeling. This is called \emph{Affective Primacy}~\citep{Zajonc80feelingand}. For example, our immediate reaction when we encounter an unknown creature in the jungle without evaluating whether it is safe or dangerous. Sensory inputs $Q$ can also trigger cognitive modeling for subsequent evaluation of the situation and thus can update the latent speaker-state $S$.

\textbf{Emotional-aware state ($E$):} Emotional-aware state encodes the emotion of the speaker at time $t$. As proposed by the psychology theorist Lazarus in his article~\citep{lazarus1982thoughts}, the emotional-state can be triggered by cognition and thinking, we think in a conversation, this state should be controlled by $S$ and $I$. If we refer to the same example above, the expressed emotion by person B depends on the speaker-state $S$ and intent $I$. According to the theory of affective primacy by \citep{Zajonc80feelingand}, the affective or emotional state does not always depend on cognition, and in various situations, an affective response can be spontaneous without relying on any prior cognitive evaluation of the situation e.g., we fear when we encounter a snake, our eyes blink when we are exposed to sudden bright light source. Based on this theory, the state $E$ may not always depend on $S$ and $I$. In the case of affective primacy, affect directly depends on the sensory or external inputs, i.e., $Q$ which are very important in multimodal conversations.


\textbf{The workflow:} At turn $t$, the speaker conceives several pragmatic concepts, such as argumentation logic, viewpoint, personality, conversational history, emotion coding sequence of the interlocutors in the conversation and inter-personal relationships---which can collectively construct the latent speaker-state $S$~\citep{hovy1987generating} by the means of cognition. Next, the intent $I$ of the speaker is formulated based on the current speaker-state, personality and previous intent of the same speaker (at $t-2$). Personality $P$, speaker-state $S$, intent $I$ and external or sensory inputs $Q$ can jointly influence the emotion $E$ of the speaker. Finally, the intent, the speaker state, and the speaker's emotion jointly manifest as the spoken utterance.

\paragraph{\textbf{Aspect-level text summarization}}

\begin{figure*}[ht!]
	\centering
	\includegraphics[width=\linewidth]{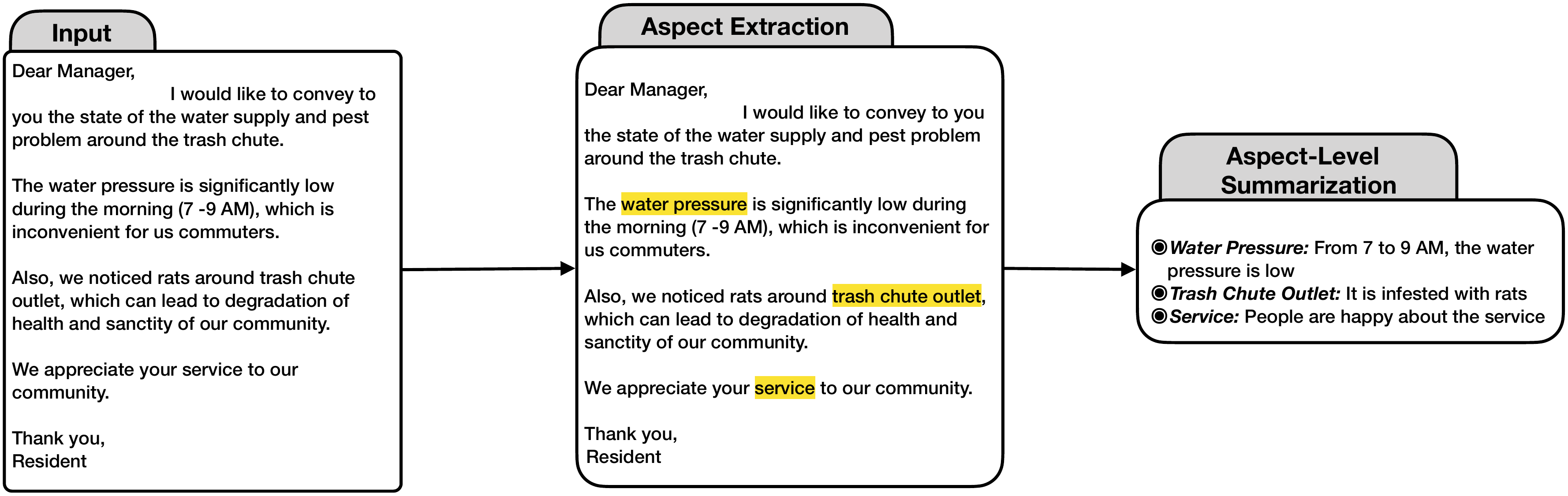}
	\caption{Aspect-level text summarization: an application of conditional generative modeling.}
	\label{fig:conditional_nlg}
\end{figure*}

One application of conditional generative models is aspect-level text summarization(~\cref{fig:conditional_nlg}). Aspect-level text summarization task consists of two stages: aspect extraction and aspect summarization. The first stage distills all the aspects discussed in the input text. The latter generates an abstractive or extractive gist of the extracted aspects from the source text. In the case of abstractive gist, the model generates content conditioned on the given aspect. Existing works~\citep{frermann2019inducing} address this task by leveraging aspect-level document segmentation and use that to generate both abstractive and extractive aspect-level summaries.

\subsubsection{Sentiment-Aware Dialogue Generation}

The area of affect-controlled text has also percolated into dialogue systems. The aim here is to equip emotional intelligence into these systems to improve user interest and engagement~\citep{partala2004effects, prendinger2005empathic}. Two key functionalities are important to achieve this goal~\citep{DBLP:conf/acl/HasegawaK0T13}:
\begin{enumerate}
    \item Given a user query, determine the best emotional/sentiment response adhering to social rules of conversations.
    \item Generate the response eliciting that emotion/sentiment.
\end{enumerate}

Present works in this field either approach these two sub-problems independently~\citep{ghosh2017affect} or in a joint manner~\citep{gu2019towards}. The proposed models range over various approaches, which include affective language models~\citep{ghosh2017affect} or seq2seq models customized to generate emotionally-conditioned text~\citep{zhou2018emotional,asghar2018affective}.
\citet{kong2019adversarial} take an adversarial approach to generate sentiment-aware responses in the dialogue setup conditioned on sentiment labels.
For a brief review of some of the recent works in this area, available corpora and evaluation metrics, please refer to~\citet{DBLP:journals/corr/abs-1906-09774}.

Despite the recent surge of interest in this application, there remains significant work to be done to achieve robust emotional dialogue models. Upon trying various emotional response generation models such as ECM~\citep{zhou2018emotional}, we surmise, these models lack the ability of conversational emotion recognition and tend to generate generic, emotionally incoherent responses. Better emotion modeling is required to improve contextual emotional understanding~\citep{hazarika2018icon}, followed by emotional anticipation strategies for the response generation. These strategies could be optimized to steer the conversation towards a particular emotion~\citep{lubis2018eliciting} or be flexible by proposing appropriate emotional categories. The quest for better text with diversity and coherence and fine-grained control over emotional intensity are still open problems for the generation stage. Also, automatic evaluation is a notorious problem that has plagued all applications of dialogue models. 

To this end, following the work by \citet{hovy1987generating}, we illustrate a sentiment and emotion-aware dialogue generation framework in Figure~\ref{fig:controlling_vars} that can be considered as the basis of future research. The model incorporates several cognitive variables, i.e., intent, sentiment, and interlocutor's latent state for coherent dialogue generation.


\subsubsection{Sentiment-Aware Style Transfer}
Style transfer of sentiment is a new area of research. It focuses on flipping the sentiment of sentences by deleting or inserting new sentiment-bearing words. E.g., to change the sentiment of \emph{"The chicken was delicious"}, we need to find a replacement of the word \emph{delicious} that carries negative sentiment. 

Recent methods on sentiment-aware style transfer attempt to disentangle sentiment bearing contents from other non-sentiment bearing parts in the text by relying on rule-based~\citep{li2018delete} and adversarial learning-based~\citep{DBLP:conf/acl/JohnMBV19} techniques. 

Adversarial learning-based methods to sentiment style transfer suffer from the lack of available parallel corpora, which opens the door to a potential future work. Some initial works, such as~\citep{DBLP:conf/nips/ShenLBJ17}, address non-parallel style transfer, albeit with strict model assumptions. We also think this research area should be studied together with the ABSA (aspect-based sentiment analysis) research to learn the association between topics/aspects and sentiment words. Considering the example above, learning better association between topics/aspects and opinionated words should aid a system to substitute \emph{delicious} with \emph{unpalatable} instead of another negative word \emph{rude}.


\subsection{Bias in Sentiment Analysis Systems}

Fairness in machine learning has gained much traction recently. Amongst multiple proposed definitions of fairness, we particularly look into ones that attempt to avoid disparate treatment and thus reduce the disparate impact across demographics. This requires removing bias that favors a subset of the population with an unfair advantage.


Studying fairness in sentiment analysis is crucial, as diverse demographics often share the derived commercial systems. Sentiment analysis systems are often used in sensitive areas, such as healthcare, which deals with sensitive topics like counseling. Customer calls and marketing leads, from various backgrounds, are often screened for sentiment cues, and the acquired analytics drives major decision-making. Thus, understanding the presence of harmful bias is critical. Unfortunately, the field is at its nascent stage and has received minimal attention. However, some developments have been observed in this area, which opens up numerous research directions. 

While there can be different demographics, such as gender, race, age, etc., our subsequent discussions, without any loss of generality, primarily exemplify gender biases. 

\subsubsection{Identifying Causes of Bias in Sentiment Analysis Systems}
Bias can be introduced into the sentiment analysis models through three main sources: 
\begin{enumerate}
    \item \textit{Bias in word embeddings}: Word embeddings are often trained on publicly available sources of text, such as Wikipedia. However, a survey by \citet{women15} found that less than 15\% of contributions to Wikipedia come from women. Therefore, the resultant word embeddings would naturally under-represent women's point of view.
    \item \textit{Bias in the model architecture}: Sentiment-analysis systems often use meta information, such as gender identifiers and indicators of demographics that include age, race, nationality, and geographical cues. Twitter sentiment analysis is one such application where conditioning on these variables is
    prevalent~\citep{DBLP:journals/corr/abs-1302-3299, DBLP:conf/wassa/VosoughiZR15, volkova2013exploring}. Though helpful, such design choices can often lead to various forms of bias such as gender bias, geographic (location) bias, etc. from theses conditioned variables. These model architectures can further amplify the bias observed in the training data. 
    
    Sentiment intensity of a word or phrase can be interpreted differently across geographical demographics. For example, the general perception about \emph{good weather}, \emph{good traffic}, \emph{cheap phone} can vary among the Indian and American populations. Hence, training a sentiment model on data originating from one geographic location can expose the model to demographic bias when applied to data from other locations. 
    Thus, depending on the end application and region of interest, a cogent solution to this issue could be to develop demographic-specific sentiment analysis models rather than creating a generic one. However, as data is often a combination of multiple demographic classes --- e.g. a combination of various ages, genders, and locations, etc. --- building mutually exclusive demographic-based models would be computationally prohibitive. Nevertheless, research in domain adaptation is one possible avenue towards solving this problem as it would help adapting a trained sentiment model on a source demographic data to learn the features of the target demography.
    
    \item \textit{Bias in the training data}: There are different scenarios where a sentiment-analysis system can inherit bias from its training data. These include highly frequent co-occurrence of a sentiment phrase with a  particular gender --- for example, \emph{woman} co-occurring with \emph{nasty} ---, over- or under-representation of a particular gender within the training samples, strong correlation between a particular demographic and sentiment label --- for instance, samples from \emph{female} subjects frequently belonging to \emph{positive} sentiment category.
\end{enumerate}

An author's stylistic sense of writing can also be one of the many sources of bias in sentiment systems. E.g., one person uses strong sentiment words to express a positive opinion but prefers to use milder sentiment words in exhibiting negative opinions. Similarly, sentiment expression might vary across races and genders, e.g., as shown in a recent study by \citet{bhardwaj2020investigating}. As a result, a sentiment analysis model might show a drastic difference in the sentiment intensities when the gender word in the same sentence is changed from masculine to feminine or vice versa, making the task of identifying bias and de-biasing difficult.  

\subsubsection{Evaluating Bias} 
Recent works present corpora that curate examples, specifically to evaluate the existence of bias. The Equity Evaluation Corpus (EEC)~\citep{DBLP:conf/starsem/KiritchenkoM18} is one such example that focuses on finding \textit{gender} and \textit{racial} bias. The sentences in this corpus are generated using simple templates, such as ``\verb|<Person>| made me feel \verb|<emotional state word>|",  ``\verb|<Person>| feels angry". The variable \verb|<Person>| can be a female name such as ``Jasmine", or a male name such as ``Alan". A pre-trained sentiment or emotion intensity predictor is then tasked to predict the emotion and sentiment intensity of the sentences. According to the task setting, a model can be called gender-biased when it consistently or significantly predicts higher or lower sentiment intensity scores for sentences carrying female-names than male-names, or vice versa. The EEC corpus contains $7$ templates of type: \verb|<Person>| and \verb|<emotional state word>|. The placeholder \verb|<Person>| can be filled by any of 60 gender-specific names or phrases. Out of those 60 gender-specific names, 40 are gender-specific names (20-female, 20-male). Rest 20 are noun phrases grouped as female-male pairs such as ``my mother" and ``my father". Variable \verb|<emotional state word>| can take four emotions -- Anger, Fear, Sadness, and Joy -- each having 5 representative words~\footnote{e.g.,:- \{angry, enraged\} represents a common emotion, anger}.
Thus, there are 1200 ($60 \times 5 \times 4$) samples for each template. Finally, there are 8400 ($7 \times 1200$) samples equally divided in female and male-specific sentences ($60 \times(5 \times 4)\times 7=4200$ each) and 4 emotion categories ($5 \times 7 \times 60=2100$ each). We refer readers to ~\cite{DBLP:conf/starsem/KiritchenkoM18} for an elaborative explanation on the EEC corpus. While this is a good step, the work is limited to exploring bias that is related only to gender and race. Moreover, the templates utilized to create the examples might be too simplistic, and identifying such biases and de-biasing them might be relatively easy. Future work should design more complex cases that cover a wider range of scenarios. Challenge appears when we have scenarios like ``\emph{John told Monica that she lost her mental stability}" vs. ``\emph{John told Peter that he lost his mental stability}". If the sentiment polarity in either of these two sentences is predicted significantly different from the other, that would indicate a likely gender bias issue.

\subsubsection{De-biasing}
If a model shows the sign of bias because of the training data, one approach to curb this bias is to distill a subset that does not display any explicit indication of bias. However, often it is unfeasible to collate data meeting such constraints. Hence, it is the responsibility of researchers to come up with solutions that can de-bias any bias introduced by the training data and model architecture.

The primary approach to de-biasing is to perturb a text with word substitution to generate counterfactual cases in the training data. These generated instances can then be used to regularize the learning of the model, either by constraining the embedding spaces to be invariant to the perturbations or minimizing the difference in predictions between both the correct and perturbed instances. While recent approaches, such as~\citep{DBLP:journals/corr/abs-1911-03064}, have proposed these methods in language models, another direction could be to mask out bias contributing terms during training. However, such a method presents its own challenges since masking might cause semantic gaps.

In general, we observe that while many works demonstrate or discuss the existence of bias, and also propose bias detection techniques, there is a shortage of works that propose de-biasing approaches. 

Existing studies mostly focus on identifying gender-bias in context-independent word representations such as GloVe~\citep{bolukbasi2016man,DBLP:conf/acl/KanekoB19,DBLP:journals/tacl/KumarBKC20}. Contrarily, BERT word-to-vector(s) mapping is highly context-dependent which makes it difficult to analyse biases intrinsic to BERT. While a lot has been studied, identified, and mitigated when it comes to gender-bias in static word embeddings~\citep{bolukbasi2016man, zhao-etal-2018-learning, caliskan2017semantics, zhao-etal-2018-gender}, very few recent works study gender-bias in contextualized settings. In \cite{zhao-etal-2019-gender, basta-etal-2019-evaluating, gonen-goldberg-2019-lipstick}, the authors apply debiasing on ELMo. \cite{kurita-etal-2019-measuring} propose a template-based approach to quantify bias in BERT. \citet{sahlgren-olsson-2019-gender} study bias in both contextualized and non-contextualized Swedish embeddings.

Apart from the traditional bias in models, bias can also exist at a higher level when making research choices. A simple example is the tendency of the community to resort to English-based corpora, primarily due to the notion of increased popularity and wider acceptance. Such trends diminish the research growth of marginalized topics and study of arguably more interesting languages -- a gap which widens through time~\citep{DBLP:conf/acl/HovyS16}. As highlighted in~\cref{sec:multilingual}, as a community, we should make conscious choices to help in the equality of under-represented communities within NLP and Sentiment Analysis.




\section{Conclusion} \label{sec:conclusion}
Sentiment analysis is often regarded as a simple classification task to categorize contents into positive, negative, and neutral sentiments. In contrast, the task of sentiment analysis is highly complex and governed by multiple variables like human motives, intents, contextual nuances. Disappointingly, these aspects of sentiment analysis remain either un- or under-explored. 

Through this paper, we strove to diverge from the idea that sentiment analysis, as a field of research, has saturated. We argued against this fallacy by highlighting several open problems spanning across
subtasks under the umbrella of sentiment analysis, such as aspect-level sentiment analysis, sarcasm analysis, multimodal sentiment analysis, sentiment-aware dialogue generation, and others. Our goal was to debunk, through examples, the common misconceptions associated with sentiment analysis and shed light on several future research directions.  We hope this work would help reinvigorate researchers and students to fall in love with this immensely interesting and exciting field, again.


\section*{Acknowledgements}
The authors are indebted to all the pioneers and researchers who contributed to this field. 

This research is supported by A*STAR under its RIE 2020 Advanced Manufacturing and Engineering (AME) programmatic grant, Award No. -  A19E2b0098, Project name - K-EMERGE: Knowledge Extraction, Modelling, and Explainable Reasoning for General Expertise.
Any opinions, findings, and conclusions or recommendations expressed in this material are those of the authors and do not necessarily reflect the views of A*STAR.

\section*{Note}
This paper will be updated periodically to keep the community abreast of any latest developments that inaugurate new future directions in sentiment analysis. The most updated version of this article can be found at \url{https://arxiv.org/pdf/2005.00357.pdf}.

\bibliographystyle{icml2019.bst}
\bibliography{icml_ref}

%
\vskip -2\baselineskip plus -1fil
\begin{IEEEbiography}[{\includegraphics[width=1in,height=1.25in,clip,keepaspectratio]{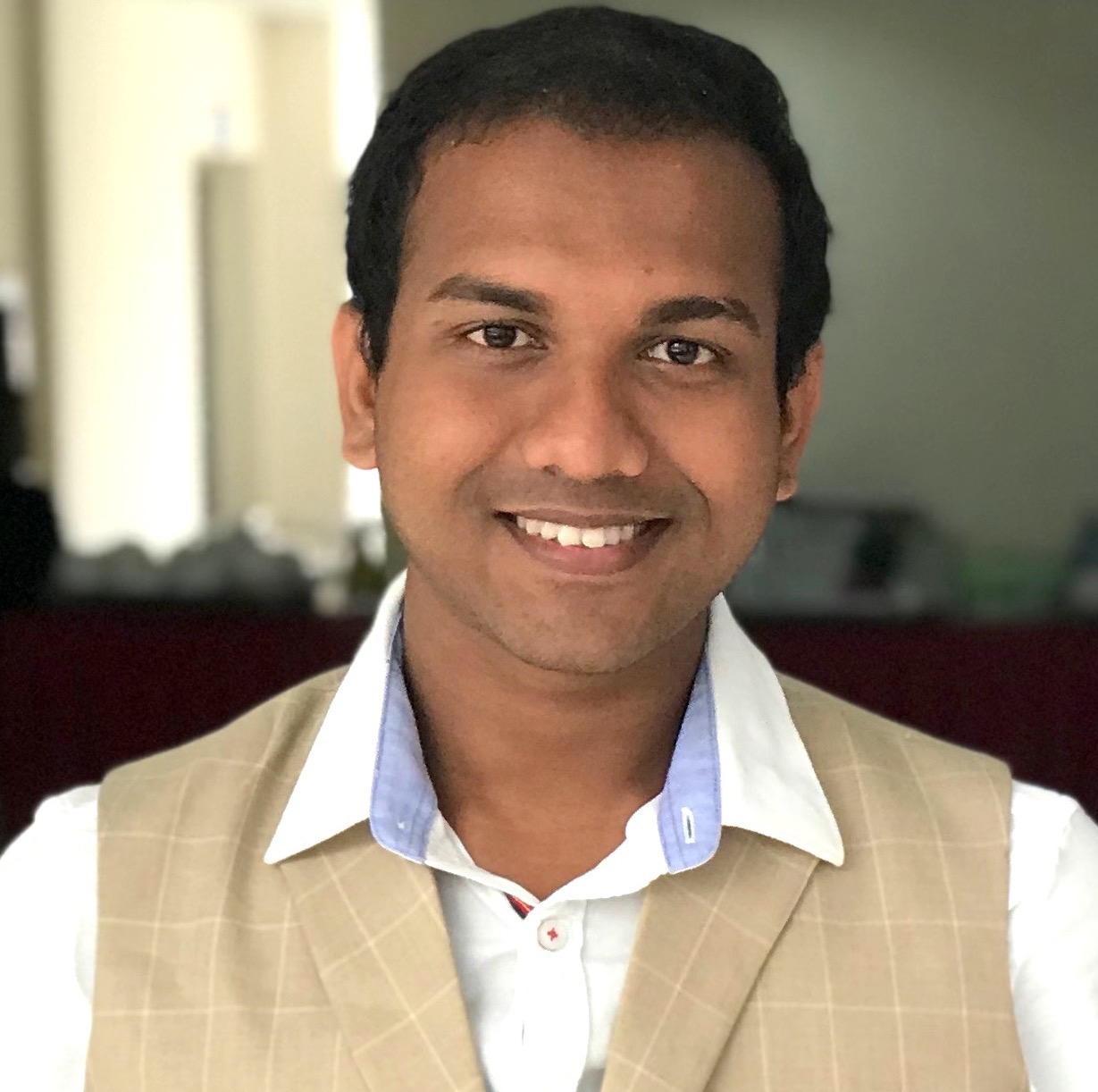}}]{Soujanya Poria} is an assistant professor of Information Systems Technology and Design, at the Singapore University of Technology and Design (SUTD), Singapore. He holds a Ph.D. degree in Computer Science from the University of Stirling, UK. He is a recipient of the prestigious early career research award called ``NTU Presidential Postdoctoral Fellowship" in 2018. Soujanya has co-authored more than 100 research papers, published in top-tier conferences and journals such as ACL, EMNLP, AAAI, NAACL, Neurocomputing, Computational Intelligence Magazine, etc. Soujanya has been an area chair at top conferences such as ACL, EMNLP, NAACL. Soujanya serves or has served on the editorial boards of the Cognitive Computation and Information Fusion.
\end{IEEEbiography}
\vskip -2\baselineskip plus -1fil
\begin{IEEEbiography}[{\includegraphics[width=1in,height=1.25in,clip,keepaspectratio]{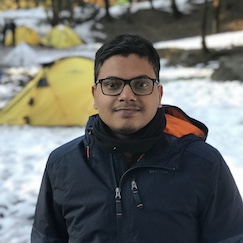}}]{Devamanyu Hazarika} is a senior Ph.D. candidate at NUS, Singapore. His primary research is on analyzing the role of context in multimodal affective systems. Devamanyu has published more than 25 research papers in top-tier conferences and journals such as ACL, EMNLP, AAAI, NAACL, Information Fusion, etc. He is also one of the prestigious presidential Ph.D. scholarship holders at NUS. He has been awarded the NUS Research Achievement Award and has been a PC-member in many top-tier conferences and journals.
\end{IEEEbiography}

\vskip -2\baselineskip plus -1fil
\begin{IEEEbiography}[{\includegraphics[width=1in,height=1.25in,clip,keepaspectratio]{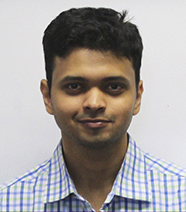}}]{Navonil Majumder} is a research fellow at the Singapore University of Technology and Design (SUTD). He has received his MSc and Ph.D. degrees from CIC-IPN in 2017 and 2020, respectively. Navonil was awarded the L\'azaro C\'ardenas Prize for his MSc and Ph.D. studies. He has also received the best Ph.D. thesis award by the Mexican Society for Artificial Intelligence (SMIA). His research interests encompass natural language processing, dialogue understanding, sentiment analysis, and multimodal language processing. Navonil has more than 25 research publications in top-tier conferences and journals, such as, ACL, EMNLP, AAAI, Knowledge-Based Systems, IEEE Intelligent Systems, etc. Navonil's research works have occasionally been featured in news portals such as Techtimes, Datanami, etc.
\end{IEEEbiography}
\vskip -2\baselineskip plus -1fil
\begin{IEEEbiography}[{\includegraphics[width=1in,height=1.25in,clip,keepaspectratio]{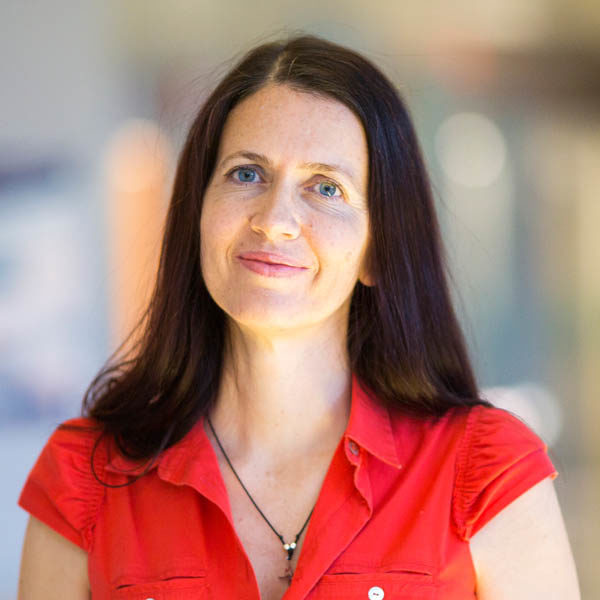}}]{Rada Mihalcea} is a Professor of Computer Science and Engineering at the University of Michigan and the Director of the Michigan Artificial Intelligence Lab. Her research interests are in lexical semantics, multilingual NLP, and computational social sciences. She serves or has served on the editorial boards of the Journals of Computational Linguistics, Language Resources and Evaluations, Natural Language Engineering, Journal of Artificial Intelligence Research, IEEE Transactions on Affective Computing, and Transactions of the Association for Computational Linguistics. She was a program co-chair for EMNLP 2009 and ACL 2011, and a general chair for NAACL 2015 and *SEM 2019. She currently serves as the ACL Vice-President Elect. She is the recipient of an NSF CAREER award (2008) and a Presidential Early Career Award for Scientists and Engineers awarded by President Obama (2009). In 2013, she was made an honorary citizen of her hometown of Cluj-Napoca, Romania.
\end{IEEEbiography}




\end{document}